\documentclass[letterpaper]{article} 
\usepackage{aaai2027}  
\usepackage[hyphens]{url} 
\usepackage{graphicx} 
\usepackage{natbib} 
\usepackage{caption} 
\usepackage{algorithm}
\usepackage{algorithmic}

\usepackage{booktabs}
\usepackage{multirow}
\usepackage{amsmath}
\usepackage{amssymb}

\usepackage{newfloat}
\usepackage{listings}

\usepackage{xcolor}
\usepackage[most]{tcolorbox}

\DeclareCaptionStyle{ruled}{labelfont=normalfont,labelsep=colon,strut=off} 

\newcounter{promptref}[section]

\floatstyle{ruled}
\newfloat{listing}{tb}{lst}{}
\floatname{listing}{Listing}
\newtcolorbox[auto counter]{promptbox}[2][]{
enhanced, float=!t, colback=black!5, colframe=black, colbacktitle=black, coltitle=white, fontupper=\normalsize, fonttitle=\bfseries\normalsize, title={Prompt~\thetcbcounter: #2}, arc=2.5mm, boxrule=0.8pt, left=2mm, right=2mm, top=1.5mm, bottom=1.5mm, #1
}
\title{Deeply Interleaved Text-Image Contexts for Multimodal LLMs Assessment}
\author{
    Zihao Wang\textsuperscript{1}, 
    Xi Xiang\textsuperscript{1},
    Yuwen Sun\textsuperscript{1},
    Yingyu Li\textsuperscript{1},
    Yabo Zhang\textsuperscript{1},
    Yihan Zeng\textsuperscript{2},
    Fan Li\textsuperscript{2,3  *},
    Wangmeng Zuo\textsuperscript{1,}
    \thanks{Corresponding authors.}
}
\affiliations{
    \textsuperscript{1}Harbin Institute of Technology, 
    \textsuperscript{2}Huawei Noah’s Ark Lab, 
    \textsuperscript{3}Nankai University
}

\begin{document}

\maketitle

\begin{abstract}
%
Current evaluations and training of multimodal models predominantly focus on multi-image tasks, largely overlooking interleaved text-image scenarios.
In such multi-image tasks, text typically serves merely as task instructions, lacking deep semantic interaction with the visual content.
In contrast, real-world applications like text-image co-creation, character tracking, and spatial reconstruction require constant interaction between text and images.
Consequently, 
models must possess a deep understanding of these interleaved contexts.
To bridge this gap, we introduce a novel benchmark, \textbf{TIC-Bench} (deeply interleaved \textbf{T}ext-\textbf{I}mage \textbf{C}ontexts), designed to evaluate the capability of models to integrate text-image clues and recover the ground truth facts within deeply interleaved contexts.
This benchmark encompasses three core domains: Logical, Temporal, and Spatial Association, which are further categorized into eight specific types, comprising a total of 2,280 questions.
We evaluated 10 state-of-the-art MLLMs and observed a substantial performance gap compared to human experts, together with persistent difficulties in integrating evidence distributed across interleaved visual and textual inputs.
Ultimately, this benchmark provides a valuable analytical tool for assessing and advancing the ability of multimodal models to effectively integrate text and image information in deeply interleaved contexts.
TIC-Bench is publicly available at \url{https://huggingface.co/datasets/pino10010/TIC-Bench}.
\end{abstract}
\begin{figure}[!t]
    \centering
    \includegraphics[width=\columnwidth]{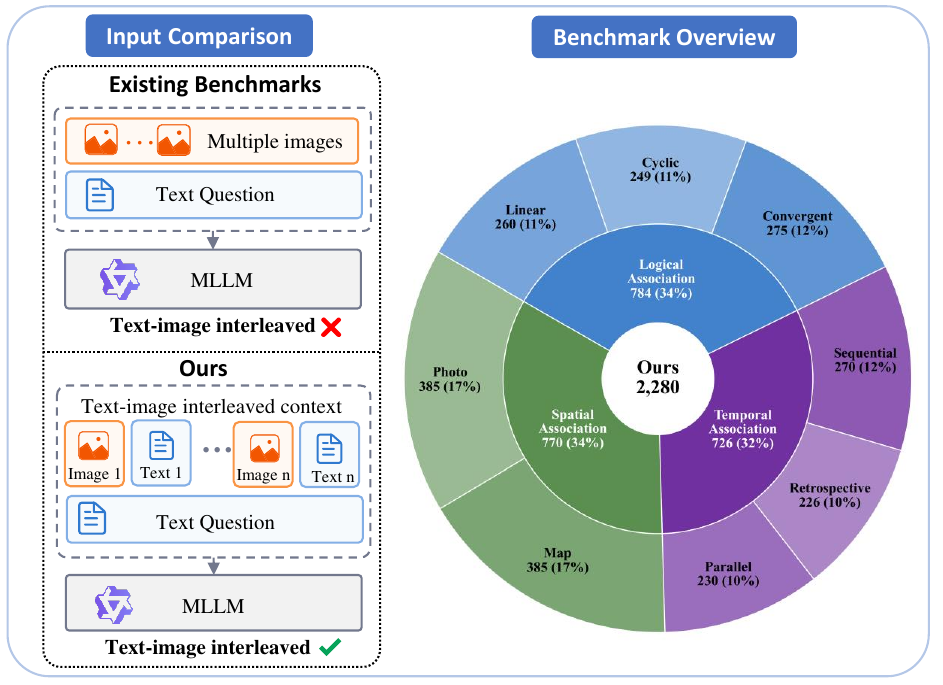}
    \caption{Overview of TIC, including its interleaved formulation  and task distribution.}
    \label{fig:teaser}
\end{figure}

\section{Introduction}

\begin{table*}[!t]
\centering
\small
\setlength{\tabcolsep}{5.0pt}
\renewcommand{\arraystretch}{1.12}
\resizebox{\textwidth}{!}{
\begin{tabular}{lccccccc}
\toprule
Benchmark
& Samples
& Total References
& References / Sample
& Images
& Images / Sample
& References / Image
& Avg. Words \\
\midrule

MMMU~\cite{yue2024mmmu}
& 11,550
& 11,821
& 1.02
& 13,281
& 1.15
& 0.89
& 38.2 \\

MMIU~\cite{meng2025mmiu}
& 11,698
& 66,973
& 5.73
& 79,259
& 6.78
& 0.84
& 127.7 \\

MIRBench~\cite{du2025easy}
& 10,397
& 68,159
& 6.56
& 43,301
& 4.16
& 1.57
& 119.6 \\

MMRB~\cite{cheng2025evaluating}
& 4,750
& 40,727
& 8.57
& 29,293
& 6.17
& 1.39
& 96.7 \\

MIRACLE~\cite{zhu2026will}
& 1,001
& 8,502
& 8.49
& 6,726
& 6.72
& 1.26
& 39.0 \\

MMR-Life~\cite{li2026mmr}
& 2,646
& 1,694
& 0.64
& 19,108
& 7.22
& 0.09
& 65.1 \\

VisReason~\cite{guo2026can}
& 1,505
& 1,278
& 0.85
& 1,605
& 1.07
& 0.80
& 33.9 \\

MIMIC~\cite{das2026more}
& 5,168
& 34,344
& 6.65
& 34,443
& 6.66
& 1.00
& 15.0 \\

\midrule

\textbf{TIC (Ours)}
& 2,280
& \textbf{85,145}
& \textbf{37.34}
& 45,776
& \textbf{20.08}
& \textbf{1.86}
& \textbf{309.3} \\

\bottomrule
\end{tabular}
}
\caption{
Comparison with existing multi-image and interleaved text--image benchmarks. 
Despite a moderate overall sample size, \textbf{TIC-Bench} features the largest total number of image references and leads across all key context complexity metrics: references per sample, images per sample, references per image, and average textual context length. 
These statistics underscore the benchmark's focus on evaluating long, densely interleaved multimodal contexts. Image counts are aggregated per sample, while image references encompass both structural image blocks and explicit in-text mentions. The highest values for context complexity in each column are highlighted in bold.
}
\label{tab:benchmark_statistics}
\end{table*}
In recent years, multimodal large language models (MLLMs)~\cite{bai2023qwen,yang2025qwen3,team2026gemma,hong2025glm,team2025kimi,team2026full} have achieved remarkable progress in multimodal understanding and reasoning.
Existing models have demonstrated substantial progress in processing single text-image pairs~\cite{li2023blip} and traditional multi-image scenarios ~\cite{suhr2019corpus, huang2016visual}.
However, applying MLLMs to complex real-world scenarios imposes significantly more stringent demands.
Typical applications, including multimodal document understanding ~\cite{yan2025m,hu2024mplug}, text-image collaborative creation~\cite{cui2025emu3,wang2026illuminating,deng2025emerging,wang2026premier,Li_2026_CVPR,Li_2026_CVPR_HPEdit,wang2025ace}, and multi-perspective event tracking~\cite{tang2019cityflow, feng2021cityflow, zhou2023joint}, inherently present information in highly interleaved formats.
Solving these tasks requires models to reason over interleaved text-image contexts rather than perceiving images and text in isolation.
An effective benchmark should therefore assess whether models can continuously bind, integrate, and propagate fragmented visual and textual clues across extended interleaved sequences.

Despite the critical need for such deep interleaved text-image reasoning, current evaluation paradigms exhibit fundamental limitations in fostering these capabilities.
From a training perspective, mainstream pre-training corpora~\cite{gadre2023datacomp,schuhmann2022laion,changpinyo2021conceptual} predominantly consist of isolated text-image pairs, depriving models of the high-density interleaved data necessary to learn how to extract and integrate cross-modal clues.
Consequently, existing models often degenerate into shallow pattern matching, heavily relying on language priors rather than engaging in genuine cross-modal reasoning across interleaved contexts ~\cite{lee2025vlind,chen2024we}.
Existing evaluation benchmarks~\cite{yue2024mmmu} provide limited coverage of this capability, as they predominantly focus on single-image question answering or multi-image comprehension with images presented as a parallel collection, rather than requiring models to follow fine-grained text-image correspondences and integrate evidence across an interleaved sequence.
Ultimately, this design reduces complex tasks to simple static perception, failing to evaluate how models process interleaved text-image inputs and perform dynamic reasoning.

To bridge this critical gap, we introduce \textbf{TIC-Bench}, a novel benchmark specifically designed to break this decoupling and compel models to reason over continuously interleaved text-image contexts.
By minimizing the extent to which questions can be answered using parametric knowledge alone, TIC-Bench emphasizes genuine cross-modal evidence integration.
As illustrated in Figure~\ref{fig:teaser},TIC-Bench encompasses three core dimensions: Logical Association, Temporal Association, and Spatial Association, which are further divided into eight task types with distinct reasoning structures.
Logical Association comprises Linear, Cyclic, and Convergent logic, requiring models to ground textual references in visual objects and follow either a single chain, a chain that revisits previous scenes, or multiple chains whose results must be integrated.
Temporal Association includes Sequential, Retrospective, and Parallel scenarios, evaluating whether models can track events and character states by following subsequent developments, retrieving evidence from earlier scenes, or coordinating multiple concurrent storylines.
Spatial Association consists of Map and Photo reasoning, requiring models to combine fragmented visual patches with interleaved textual descriptions to infer relative positions in map-based or natural-image environments.
Together, these tasks evaluate models' ability to maintain cross-modal correspondences and integrate distributed evidence across diverse interleaved contexts.

%
Through extensive evaluations on TIC-Bench, we identify several systematic performance patterns and recurring failure modes of current MLLMs.
First, a substantial gap remains between state-of-the-art models and human experts.
The strongest evaluated model, GPT-5.5, achieves an overall accuracy of 59.9\%, compared with 91.7\% for the human baseline.
Second, model performance varies considerably across reasoning structures.
Within the Logical Association domain, models generally perform better on Cyclic and Linear logic than on Convergent logic.
This result suggests that integrating the outcomes of multiple reasoning branches presents a more persistent challenge.
A similar pattern emerges in the Temporal Association domain, where Parallel reasoning is the most difficult temporal category for most evaluated models.
These results indicate that maintaining and merging concurrent evidence streams remains a major bottleneck for current MLLMs.
Finally, our error analysis identifies abstraction and reasoning errors as major sources of failure, particularly in maintaining consistent entity mappings and integrating evidence distributed across multiple images and textual segments.
Together, these findings demonstrate that the principal challenge of deeply interleaved multimodal reasoning lies not only in processing long contexts but also in coordinating, maintaining, and merging multiple cross-modal evidence streams.
%
The main contributions of this work are summarized as follows:
\begin{itemize}
    \item We introduce \textbf{TIC-Bench}, a novel benchmark for evaluating multimodal reasoning over deeply interleaved text-image contexts. TIC-Bench contains 2,280 questions across three complementary domains (Logical, Temporal, and Spatial Association) and eight task types with distinct reasoning structures.
    \item We conduct a comprehensive evaluation of 10 state-of-the-art MLLMs across 15 inference settings, including \textit{thinking} and standard variants for five open-source models, and compare their performance against a human baseline. The results reveal a substantial gap between the strongest evaluated model and human experts.
    \item Through detailed task-level and error-level analyses, we identify recurring limitations in coordinating multiple cross-modal evidence streams. Tasks involving Convergent logical and Parallel temporal scenarios are particularly challenging, while abstraction and reasoning errors reveal persistent difficulties in maintaining consistent entity mappings and integrating evidence distributed across multiple images and textual segments.
\end{itemize}
\begin{figure*}[t]
    \centering
    \includegraphics[width=\linewidth]{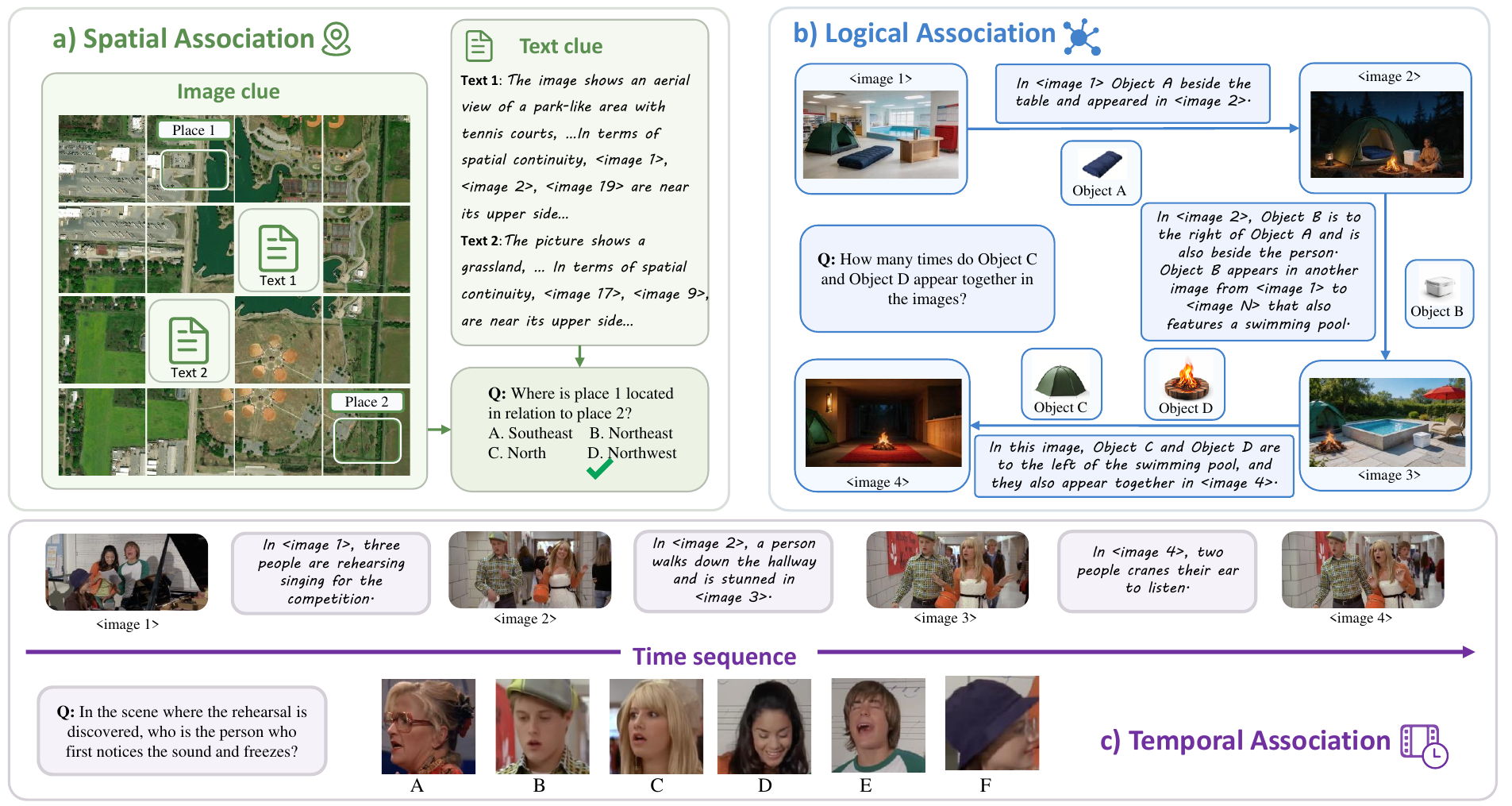}
    \caption{
    Overview of our interleaved text-image benchmark, covering spatial, logical, and temporal association reasoning.
    }
    \label{fig:dataset}
\end{figure*}

\section{Related Work}
\subsection{Multimodal Large Language Models}

Recent years have witnessed remarkable progress in Multimodal Large Language Models (MLLMs) across tasks such as visual question answering, image captioning, and visual reasoning.
However, whether existing models truly reason based on visual evidence remains an open question.
VisReason~\cite{guo2026can} proposes a vision-centric reasoning benchmark, revealing that current MLLMs still rely heavily on language priors rather than genuinely grounding their reasoning in visual evidence.
Such findings pose even greater challenges for interleaved  text-image scenarios, where deep cross-modal reasoning between images and text is required, rather than relying on clues from a single modality.

\subsection{Multimodal Evaluation Benchmarks}

%
MMMU~\cite{yue2024mmmu} collects 11.5K college-level multimodal questions spanning six disciplines, yet predominantly features single-image questions without involving relational reasoning across multiple images.
MMIU~\cite{meng2025mmiu} is the first to systematically propose multi-image understanding evaluation.
MMRB~\cite{cheng2025evaluating} evaluates spatial, temporal, and semantic reasoning across multiple images.
However, in these benchmarks, text merely serves as question descriptions or instructions, without forming deeply interleaved contexts with images.
MIRACLE~\cite{zhu2026will} designs multi-image reasoning questions emphasizing strong inter-image dependencies.
MMR-Life~\cite{li2026mmr} constructs multiple-choice questions based on real-life scenarios, covering 7 reasoning types.
MIMIC~\cite{das2026more} diagnoses the failure modes of multi-image VLMs, revealing that models struggle to aggregate information across images and simultaneously track multiple concepts.
The work most closely related to ours is MIR~\cite{du2025easy}, which explicitly introduces the concept of Multi-image Interleaved Reasoning without systematically categorize different types of associations within interleaved  text-image contexts.
Compared with these works, TIC-Bench systematically evaluates logical, temporal, and spatial associations in interleaved text-image contexts while introducing denser cross-modal references and longer multimodal contexts.
Table~\ref{tab:benchmark_statistics} summarizes these differences, showing that TIC-Bench contains substantially denser image references, more images per instance, and longer textual contexts than existing multimodal benchmarks.

\begin{figure*}[th]
    \centering
    \includegraphics[width=\linewidth]{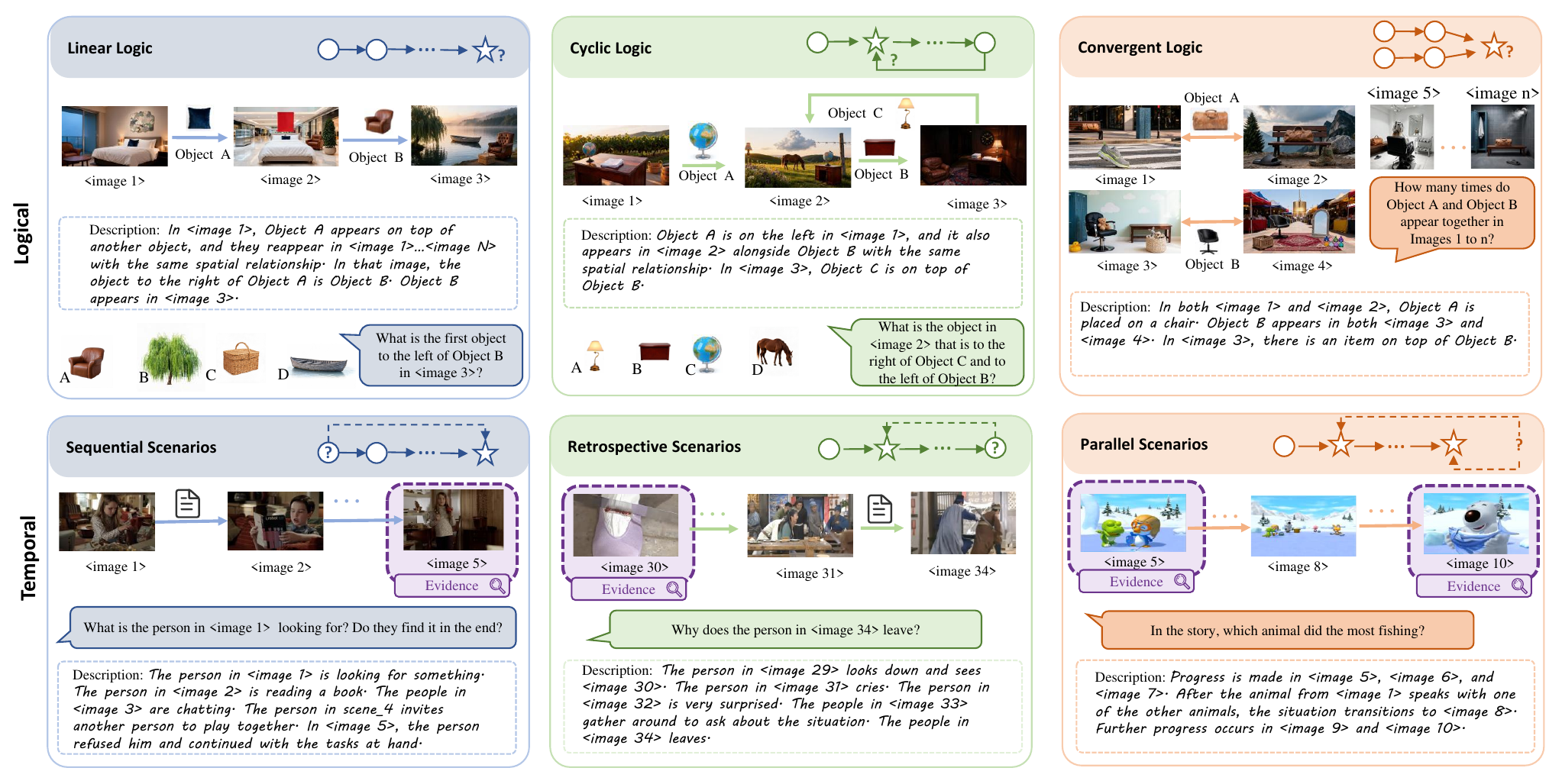}
   \caption{
Examples of logical and temporal association tasks. Logical association includes linear, cyclic, and convergent reasoning structures, while temporal association covers sequential, retrospective, and parallel scenarios.
}
\label{fig:category}
\end{figure*}

\section{ Benchmark}


\subsection{Task Definition}
In \textbf{TIC-Bench}, we formulate interleaved text-image reasoning as joint cross-modal reasoning over an interleaved multimodal context.
Specifically, each task instance consists of a context sequence $\mathcal{C}=[m_1,m_2,\dots,m_N]$,
where each element is either visual or textual, i.e., $m_i \in \mathcal{X}_{\mathcal{V}} \cup \mathcal{X}_{\mathcal{T}}$.
We use  $\tau(m_i)\in\{\mathcal{V},\mathcal{T}\}$ to denote the modality of $m_i$. The sequence does not require strict alternation between modalities; instead, it may contain consecutive elements from the same modality while maintaining multiple modality transitions.
Given the multimodal context $\mathcal{C}$ and a target query $q$, the objective of a model $f_\theta$ is to predict the answer $a=f_\theta(\mathcal{C},q)$.

Unlike traditional visual question answering tasks, each TIC-Bench instance requires complementary evidence from both the textual subset
$\mathcal{C}_{\mathcal{T}}=\{m_i\in\mathcal{C}\mid\tau(m_i)=\mathcal{T}\}$
and the visual subset
$\mathcal{C}_{\mathcal{V}}=\{m_i\in\mathcal{C}\mid\tau(m_i)=\mathcal{V}\}$,
rather than information from either modality alone.
Therefore, solving these tasks requires models to align dispersed visual and textual clues, preserve cross-modal correspondence, and integrate distributed evidence through multi-step reasoning.

\subsection{Dataset Composition and Construction}
\begin{table*}[!t]
\centering
\small
\setlength{\tabcolsep}{4.0pt}
\renewcommand{\arraystretch}{1.12}
\resizebox{\textwidth}{!}{
\begin{tabular}{l c ccc ccc cc}
\toprule
\multirow{2}{*}{Model}
& \multirow{2}{*}{Overall}
& \multicolumn{3}{c}{Logical Association}
& \multicolumn{3}{c}{Temporal Association}
& \multicolumn{2}{c}{Spatial Association} \\
\cmidrule(lr){3-5}
\cmidrule(lr){6-8}
\cmidrule(lr){9-10}
&
& Linear & Cyclic & Convergent
& Sequential & Retrospective & Parallel
& Map & Photo \\
\midrule

Human Expert
& 0.917
& 0.936 & 0.917 & 0.909
& 0.903 & 0.906 & 0.916
& 0.905 & 0.937 \\

\midrule
\multicolumn{10}{c}{\textit{Open-source models}} \\

Gemma-4-31B ~\cite{team2026gemma}
& 0.405
& 0.531 & 0.655 & 0.502
& 0.331 & 0.346 & 0.358
& 0.356 & 0.265 \\

Gemma-4-31B-thinking
& 0.430
& \underline{0.589} & \underline{0.707} & 0.520
& 0.323 & 0.381 & 0.377
& 0.369 & 0.291 \\

GLM-4.6V ~\cite{hong2025glm}
& 0.337
& 0.550 & 0.677 & 0.353
& 0.225 & 0.233 & 0.253
& 0.223 & 0.286 \\

GLM-4.6V-thinking
& 0.368
& 0.523 & 0.590 & 0.436
& 0.331 & 0.335 & 0.333
& 0.221 & 0.294 \\

Qwen3.6-35B-A3B ~\cite{yang2025qwen3}
& 0.402
& \underline{0.589} & \textbf{0.735} & 0.484
& 0.414 & 0.363 & 0.285
& 0.281 & 0.223 \\

Qwen3.6-35B-A3B-thinking
& \underline{0.470}
& 0.561 & 0.619 & \underline{0.553}
& \underline{0.542} & \underline{0.523} & 0.312
& \textbf{0.442} & 0.299 \\

Kimi-K2.6 ~\cite{team2025kimi}
& 0.448
& 0.519 & 0.676 & 0.513
& 0.521 & 0.511 & \underline{0.396}
& \underline{0.377} & \underline{0.335} \\

Kimi-K2.6-thinking
& \textbf{0.474}
& 0.586 & 0.666 & \textbf{0.555}
& \textbf{0.610} & \textbf{0.558} & \textbf{0.409}
& 0.296 & \textbf{0.343} \\

MiMo-V2.5 ~\cite{team2026full}
& 0.380
& \textbf{0.600} & \underline{0.707} & 0.375
& 0.407 & 0.420 & 0.256
& 0.229 & 0.216 \\

MiMo-V2.5-thinking
& 0.430
& 0.585 & 0.651 & 0.527
& 0.479 & 0.429 & 0.287
& 0.340 & 0.262 \\

\midrule
\multicolumn{10}{c}{\textit{Closed-source models}} \\

Claude Sonnet 4.6
& 0.474
& 0.620 & \textbf{0.733} & 0.549
& 0.511 & 0.529 & 0.296
& 0.318 & 0.368 \\

Gemini 3.1 Pro Preview
& \underline{0.590}
& \textbf{0.631} & \underline{0.715} & 0.560
& \underline{0.671} & \underline{0.668} & \textbf{0.554}
& \textbf{0.525} & \underline{0.486} \\

GPT-5.5
& \textbf{0.599}
& \underline{0.627} & 0.703 & \textbf{0.586}
& \textbf{0.685} & \textbf{0.678} & \underline{0.530}
& \underline{0.447} & \textbf{0.610} \\

Doubao-Seed-2.0-Pro
& 0.503
& 0.600 & 0.622 & \underline{0.584}
& 0.592 & 0.648 & 0.461
& 0.361 & 0.317 \\

GLM-5V-Turbo
& 0.407
& 0.515 & 0.682 & 0.521
& 0.411 & 0.361 & 0.341
& 0.260 & 0.294 \\

\bottomrule
\end{tabular}
}
\caption{
Main results on our benchmark.
Open-source models are evaluated under both non-thinking and thinking modes, denoted by the model name without or with the ``-thinking'' suffix, respectively.The best and second-best results within each model group are shown in bold and underlined, respectively.
}
\label{tab:main_results}
\end{table*}
As shown in Figure~\ref{fig:dataset}, TIC-Bench comprises three complementary subsets, which construct interleaved  text-image evaluation scenarios from three dimensions: Logical Association, Spatial Association and Temporal Association.
%

\subsubsection{Logical Association}
The Logical Association subset contains sequences of scenes connected by shared objects.
Instead of naming a target object directly, each question identifies it through a chain of cross-image relations.
For example, a narrative may refer to ``an object beside the table that also appears in the camping scene.'' It then relates this object to others through containment, proximity, or co-occurrence, extending the reasoning chain across scenes.
Solving the question requires grounding implicit textual references in the corresponding images and tracking the target object to the end of the chain.

As illustrated in Figure~\ref{fig:category}, Logical Association comprises three reasoning structures.
Linear Logic follows a single cross-image chain, Cyclic Logic revisits a previously accessed image during an intermediate reasoning step, and Convergent Logic integrates the results of multiple independent branches.

\subsubsection{Temporal Association}

Temporal Association comprises comic-style stories constructed from movie, animation, and TV-series scenes with interleaved textual descriptions.
Images and text provide complementary information: images convey character identities, scene states, and event details, while text describes actions, causal relations, and plot progression.
To prevent text-only shortcuts, key character and event information is masked or generalized, such as by replacing character names with broad references like ``person.''
Consequently, models must integrate visual identity cues with textual plot information to track characters and events over time.

As illustrated in Figure~\ref{fig:category}, this subset is divided into Sequential, Retrospective, and Parallel Scenarios. Sequential Scenarios require following the story timeline to identify subsequent events or character states. Retrospective Scenarios require retrieving evidence from earlier parts of the story. Parallel Scenarios involve multiple concurrent or independent storylines whose results must be integrated to answer the final question.

\subsubsection{Spatial Association}
Spatial Association is constructed by dividing a large image into multiple overlapping patches. 
Some patches are removed and replaced with textual descriptions of both their visual content and spatial relations to neighboring patches, creating an interleaved text-image context. 
Questions select objects or regions from visible patches, textual descriptions of missing patches, or entire patches, and ask about their relative spatial positions. 
Models must combine the visible content, descriptions of missing regions, and overlap relations among patches to locate the targets.

Based on the source images, this subset is divided into Map and Photo reasoning. Map reasoning primarily uses aerial images, emphasizing relations among roads, buildings, terrain, and other large-scale structures. Photo reasoning uses natural scenes and everyday photographs, focusing on spatial relations among objects, regions, and local visual content.

Further details of the dataset construction pipeline are provided in the supplementary material.

\subsection{Benchmarks Statistics}
The entire dataset contains 2,280 questions and 45,776 image instances in total. On average, each question contains 20.08 images, and each image is referenced 1.86 times.
The logical association subset contains 784 questions in total. Among them, 260 questions belong to linear logic, 275 to convergent logic, and 249 to cyclic logic.
The temporal association subset contains 726 questions in total. Among them, 270 involve sequential scenarios, 226 involve retrospective scenarios, and 230 involve parallel scenarios.
The spatial association subset contains 770 questions in total, including 385 map-type questions and 385 photo-type questions.
Across the three domains of TIC-Bench, Logical and Spatial Association use multiple-choice questions, whereas Temporal Association includes 418 open-ended questions.

\subsection{Comparisons with Existing Benchmarks}

As shown in Table~\ref{tab:benchmark_statistics}, our dataset comprises 2,280 samples, with an average of 37.34 image references per sample---roughly $4.4\times$ that of the next-best benchmark (MMRB, 8.57). The average number of images per sample is 20.08, approximately $2.8\times$ that of the runner-up (MMR-Life, 7.22). Regarding reference density, the references-per-image ratio reaches 1.86, the highest among all benchmarks, indicating that images are frequently reused throughout the questions. Moreover, the average textual-context length reaches 309.3 words, about $2.4\times$ that of the next-highest benchmark (MMIU, 127.7). These metrics demonstrate that our dataset features substantially higher text-image interleaving frequency and longer multimodal contexts, enabling a more thorough evaluation of models' cross-modal integration and multi-step reasoning capabilities under densely interleaved text-image conditions.

\section{Experiments}
\subsection{Experimental Setup}
We evaluate five open-source MLLMs, including Gemma-4-31B, GLM-4.6V, Qwen3.6-35B-A3B, Kimi-K2.6, and MiMo-V2.5, as well as five closed-source MLLMs, including Claude Sonnet 4.6, Gemini 3.1 Pro Preview, GPT-5.5, Doubao-Seed-2.0-Pro, and GLM-5V-Turbo.
For open-source models, we evaluate their performance under both thinking and non-thinking modes. For closed-source models, we use their default inference settings.
In addition, we include human expert performance as a reference to illustrate the gap between current models and human capability on our benchmark.
The human evaluation protocol is detailed in supplementary material.

To assess the correctness of model responses, we employ Qwen-3.7 Plus, DeepSeek V4 Pro, and Claude Opus 4.8 as automatic evaluators.
For samples where the three evaluators produce inconsistent judgments, we conduct human evaluation to ensure the reliability of the final assessment.
All models are evaluated using a unified task prompt and consistent input formatting within each benchmark domain.
For models supporting both thinking and non-thinking modes, we follow the officially supported inference controls to enable or disable explicit reasoning.
Detailed experimental settings are provided in the supplementary material.
\begin{figure}[!t]
    \centering
    \includegraphics[width=\linewidth]{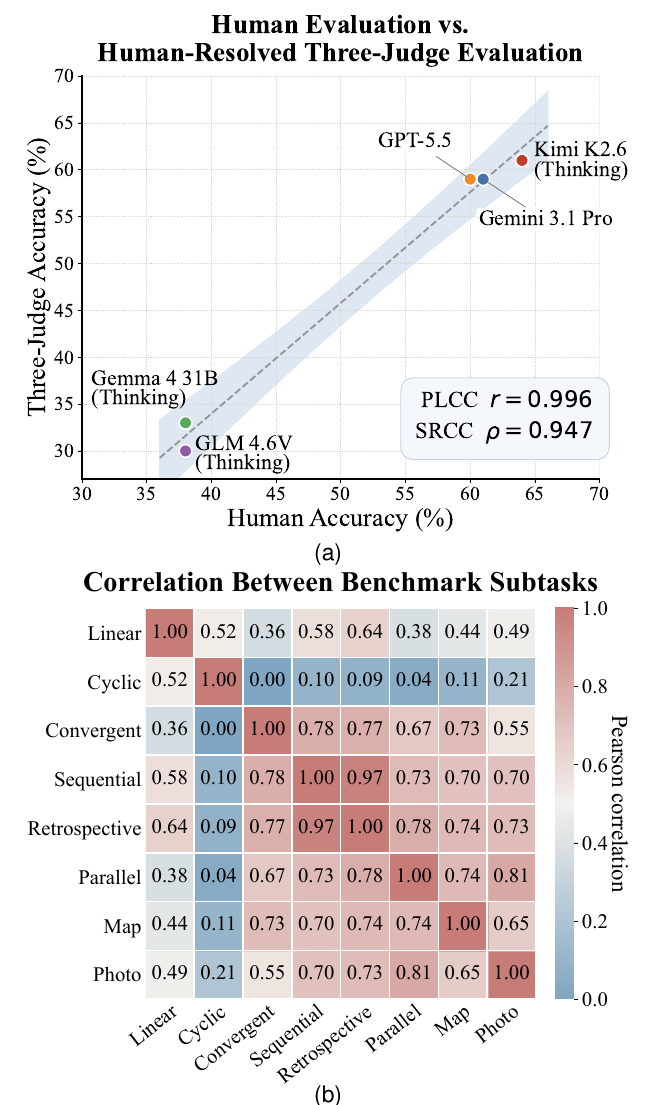}
    \caption{
    (a) Agreement between human evaluation and the human-resolved three-judge evaluation.
    (b) Pearson correlation of model performance across eight benchmark subtasks.
    }
    \label{fig:agreement_and_correlation}
\end{figure}

\subsection{Main results}

\noindent\textbf{Overall human--model gap.}
Table~\ref{tab:main_results} presents the overall performance of closed-source and open-source MLLMs on our benchmark.
The results reveal a substantial gap between current models and human experts, demonstrating that the benchmark effectively evaluates the ability to process interleaved text-image contexts.
Human experts achieve an overall accuracy of 91.7\%, whereas the strongest models, GPT-5.5 and Gemini 3.1 Pro, reach 59.9\% and 59.0\%, respectively.
This gap of more than 30 percentage points indicates that current models still struggle to sustain reasoning across distributed visual and textual evidence.
Compared with the consistently strong human performance across task types, model performance varies substantially across association types, indicating deficiencies in reasoning over specific task categories.

\noindent\textbf{Complementary strengths of closed-source models.}
Closed-source models generally outperform open-source models, although their strengths are complementary.
GPT-5.5 performs particularly well on event ordering, retrospective reasoning, and spatial reasoning over natural images, achieving 68.5\%, 67.8\%, and 61.0\% on Sequential, Retrospective, and Photo tasks, respectively.
Gemini 3.1 Pro is stronger on parallel events and map-based spatial relations, reaching 55.4\% on Parallel reasoning and 52.5\% on Map reasoning.
Claude Sonnet 4.6 obtains 73.3\% on Cyclic reasoning, further showing that no single model dominates every task category.

\noindent\textbf{Leading open-source models.}
Among open-source models, Kimi-K2.6-thinking and Qwen3.6-35B-A3B-thinking form the leading group, with overall accuracies of 47.4\% and 47.0\%, respectively.
Kimi-K2.6-thinking achieves 61.0\% on Sequential reasoning and 55.8\% on Retrospective reasoning, demonstrating a relative strength in temporal inference.
Qwen3.6-35B-A3B-thinking reaches 61.9\% on Cyclic reasoning and 44.2\% on Map reasoning, indicating stronger performance on cyclic logical structures and abstract spatial representations.

\noindent\textbf{Impact of thinking mode.}
Thinking mode generally improves open-source model performance.
For example, Qwen3.6-35B-A3B improves from 40.2\% to 47.0\% overall, while MiMo-V2.5 improves from 38.0\% to 43.0\%.
However, these gains are not consistent across all subtasks.
Explicit reasoning can help organize and extend inference chains, but it does not automatically eliminate visual recognition errors, irrelevant-context interference, or incorrect associations among multiple evidence chains.

\noindent\textbf{Challenges across task structures and modalities.}
Convergent logical reasoning and Parallel temporal reasoning remain particularly challenging.
Both require a model to maintain multiple information streams and subsequently merge or compare them.
Relative to sequential inference along a single evidence chain, such multi-branch integration is more susceptible to omissions, confusion, and incorrect entity binding.
The spatial results also exhibit a clear modality effect: most open-source models perform worse on Photo than on Map tasks, suggesting that visual clutter, object-recognition errors, and scale variation further increase the difficulty of spatial reasoning.

\noindent\textbf{Correlation between tasks.}
%
As shown in Figure~\ref{fig:agreement_and_correlation}(b), the task correlation analysis reveals capabilities shared across tasks as well as capabilities specific to individual tasks.
Sequential and Retrospective performance is highly correlated ($r=0.97$), reflecting their shared reliance on temporal order modeling.
More notably, Parallel temporal reasoning is strongly correlated with both spatial subtasks.
The correlation between Parallel and Photo is the strongest among subtask pairs from different domains ($r=0.81$), while the correlation between Parallel and Map is also strong ($r=0.74$).
This pattern suggests that coordinating concurrent event streams and reconstructing spatial relations from fragmented visual and textual observations share a common bottleneck, namely the need to maintain and integrate evidence distributed across multiple multimodal inputs.
%
In contrast, Cyclic reasoning shows weak correlations with most other subtasks, indicating that its reasoning requirements are relatively distinct.
Overall, these results identify temporal modeling and parallel multimodal reasoning as two key capabilities for effectively processing deeply interleaved text-image contexts.

\noindent\textbf{Need for stronger multimodal context management.}
Current models require improvements not only in perception and multi-step inference but also in multimodal context management.
A capable model must preserve key entities, scene states, and event relations throughout long interleaved text-image inputs, suppress irrelevant information, and maintain stable and consistent correspondences across multiple sources of visual and textual evidence.

\subsection{Modality Ablation}

%
We conduct modality ablations on four representative models Gemma-4-31B-thinking, Qwen3.6-35B-A3B-thinking, GPT-5.5, and Gemini 3.1 Pro Preview under three settings: Full, Text-only, and Images-only.
Full retains the original interleaved context, whereas Text-only and Images-only remove the visual and textual evidence, respectively.
As shown in Table~\ref{tab:modality_ablation}, Full achieves the highest overall accuracy for every model, while removing either modality causes a substantial performance drop.
For example, GPT-5.5 and Gemini 3.1 Pro Preview decrease from 59.9\% and 59.0\% under Full to 31.0\% and 31.4\% under Text-only, respectively.
Although Images-only consistently outperforms Text-only, it remains substantially below Full, demonstrating that TIC-Bench cannot be solved through single-modality shortcuts and requires the integration of complementary visual and textual evidence.

\subsection{Human Agreement Evaluation}

%
Time QA contains 418 open-ended questions. We randomly sample 100 responses from each of five representative models, yielding 500 human-scored responses, with scores of at least 80 considered correct.
Automatic evaluation uses three independent judges, namely Qwen 3.7 Plus, DeepSeek V4 Pro, and Claude Opus 4.8, with human adjudication when their judgments disagree.
This pipeline achieves 88.6\% sample-level agreement with human evaluation and a Cohen's $\kappa$ of 0.772.
The regression analysis in Figure~\ref{fig:agreement_and_correlation}(a) further shows strong agreement at the model-accuracy level, with a PLCC of 0.996 and an SRCC of 0.947, indicating that the evaluation pipeline closely approximates human-assessed accuracy while largely preserving model rankings.
\begin{table}[t]
\centering
\small
\setlength{\tabcolsep}{3.0pt}
\renewcommand{\arraystretch}{1.12}
\resizebox{\linewidth}{!}{
\begin{tabular}{llcccc}
\toprule
Model
& Input
& \shortstack{Logical\\Association}
& \shortstack{Temporal\\Association}
& \shortstack{Spatial\\Association}
& Overall \\
\midrule
\multirow{3}{*}{\shortstack[l]{Gemma-4-31B-thinking\\\cite{team2026gemma}}}
& Full        & \textbf{0.602} & \textbf{0.358} & 0.330 & \textbf{0.430} \\
& Text-only   & 0.337 & 0.221 & \textbf{0.344} & 0.301 \\
& Images-only & 0.390 & 0.270 & 0.321 & 0.327 \\
\midrule

\multirow{3}{*}{\shortstack[l]{Qwen3.6-35B-A3B-thinking\\\cite{yang2025qwen3}}}
& Full        & \textbf{0.577} & \textbf{0.463} & \textbf{0.371} & \textbf{0.470} \\
& Text-only   & 0.366 & 0.229 & 0.345 & 0.313 \\
& Images-only & 0.424 & 0.330 & 0.332 & 0.362 \\
\midrule

\multirow{3}{*}{GPT-5.5}
& Full        & \textbf{0.637} & \textbf{0.634} & \textbf{0.529} & \textbf{0.599} \\
& Text-only   & 0.302 & 0.256 & 0.373 & 0.310 \\
& Images-only & 0.485 & 0.534 & 0.320 & 0.446 \\
\midrule

\multirow{3}{*}{Gemini 3.1 Pro Preview}
& Full        & \textbf{0.633} & \textbf{0.633} & \textbf{0.506} & \textbf{0.590} \\
& Text-only   & 0.310 & 0.240 & 0.392 & 0.314 \\
& Images-only & 0.336 & 0.501 & 0.428 & 0.422 \\
\bottomrule
\end{tabular}
}
\caption{
Modality ablation results. The best input setting for each model and metric is shown in bold.
}
\label{tab:modality_ablation}
\end{table}

\begin{figure}[!t]
    \centering
    \includegraphics[width=\linewidth]{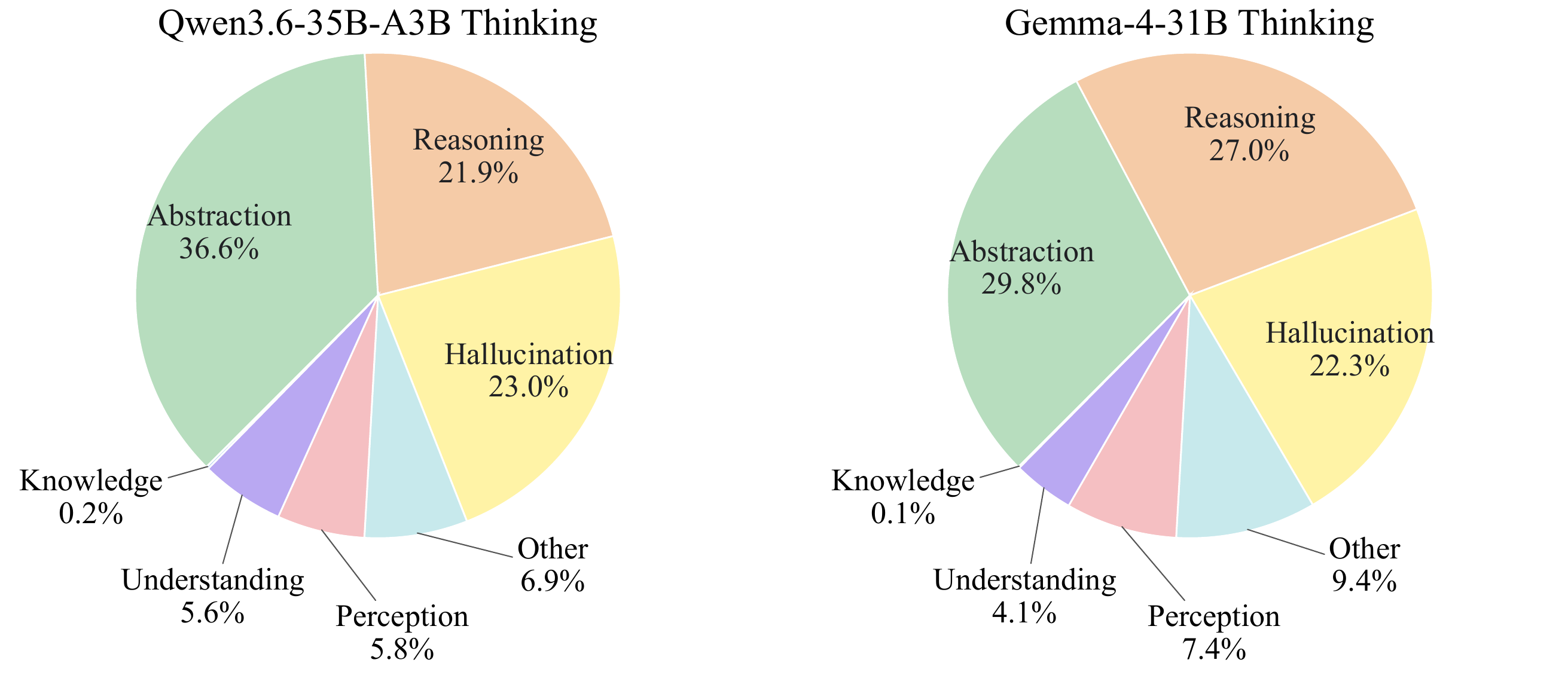}
    \caption{Distribution of primary error types for Qwen3.6-35B-A3B and Gemma-4-31B.}
    \label{fig:error_analysis}
\end{figure}
\subsection{Error Analysis}

%
We analyze Qwen3.6-35B-A3B-thinking and Gemma-4-31B-thinking because their relatively complete reasoning outputs facilitate identification of the earliest error stage.
Each incorrect response is assigned one of seven mutually exclusive labels: perception, abstraction, reasoning, understanding, hallucination, knowledge, or other.
As shown in Figure~\ref{fig:error_analysis}, Qwen is primarily affected by abstraction errors (36.6\%), followed by hallucination (23.0\%) and reasoning errors (21.9\%).
Gemma exhibits a more balanced distribution, with abstraction, reasoning, and hallucination errors accounting for 29.8\%, 27.0\%, and 22.3\%, respectively.
Abstraction errors mainly reflect inconsistent mappings between entities and symbolic references, while reasoning and hallucination errors involve failed multi-step relation composition or the introduction of unsupported information.
Knowledge errors account for less than 0.2\% for both models, indicating that the main challenges lie in entity binding, cross-scene tracking, and multi-step reasoning rather than insufficient external knowledge.
This finding also provides indirect evidence that TIC-Bench places limited demands on external prior knowledge and primarily evaluates models' ability to reason over the provided interleaved context.

\section{Conclusion}

We introduce a multimodal benchmark for association reasoning over interleaved text-image contexts, covering three domains and eight subtasks across logical, temporal, and spatial associations. Our results reveal a substantial gap between current models and human experts, with different models exhibiting complementary strengths across tasks. Although thinking mode improves performance in some cases, existing models still struggle to reliably connect evidence distributed across multiple images and text segments. Future models therefore need stronger capabilities for managing long interleaved multimodal contexts, preserving relevant information, suppressing distractions, and establishing consistent cross-modal associations.

\bibliography{aaai2027}
\clearpage
\appendix

\suppressfloats[t]

\section*{Supplementary Material Overview}

\par\noindent\textbf{Detailed Dataset Construction.}
Describes the data sources and construction pipeline for the three benchmark domains.\par
\par\noindent\textbf{Prompts.}
Collects all prompts used for dataset construction, Temporal inference, and automatic evaluation.\par
\par\noindent\textbf{Human Evaluation Protocol.}
Documents evaluator backgrounds, answering conditions, and accuracy computation.\par
\par\noindent\textbf{Model Inference and Evaluation Details.}
Reports input assembly, image processing, output parsing, and judge settings.\par
\par\noindent\textbf{Additional Analysis Protocols.}
Presents the error taxonomy and three complete representative error cases.\par
\par\noindent\textbf{Representative Benchmark Examples.}
Provides one complete text-image example for each of the eight task types.\par

\section{Detailed Dataset Construction}

\subsection{Logical Association}
\par\noindent\textbf{Structured scene and object generation.} The pipeline first samples a sequence of scene types and generates one structured visual node for each scene.
%
Qwen 3.7 Plus generates the object names, object attributes, and scene content.
%
Claude Opus 4.7 then organizes this information into a structured visual-node description that satisfies the required field constraints.
%
Each node defines 5--10 objects and records the visual attributes and image-generation description of every object.
%
The semantic attributes include category, color, size, shape, material, state, texture, temperature, position, pose, and interaction.
%
The node also describes the overall scene, the spatial and action relations among objects, and its connections to earlier scenes.
%
When the same object or relation appears across scenes, its identity and visual characteristics are kept consistent.
\par\noindent\textbf{Object and relation reference generation.} Before rendering a complete scene, GPT Image 2 generates a white-background reference image for each object from its object-level generation prompt.
%
The same model generates spatial-relation and action-relation reference images from the corresponding relation prompts.
%
These isolated references convert the object descriptions, attributes, and relations into concrete visual conditions for the final scene generator.
%
\par\noindent\textbf{Final node rendering.} The final node image is generated by Doubao Seedream 5.0 .
%
When generating the final scene, the model jointly uses all object references, relation references, and the complete scene description.
%
The object references preserve consistent appearances, while the scene description determines the environment, layout, lighting, and composition.
%
Prompt~\ref{prompt:logical_scene_generation} specifies the structured requirements for Logical scene generation, and an output is retained only after checks for field completeness and visual consistency.

Prompt~\ref{prompt:logical_question_generation} converts the question structure precomputed by code into a natural-language question.

\par\noindent\textbf{Reasoning structures.} After image generation, the code connects scenes through shared objects, attributes, and relations and uses these connections to construct three reasoning tasks.
%
Linear Logic starts from one clue and follows a single chain from image to image until the target is identified.
%
The chain does not split into multiple branches and does not require an intentional return to a previously visited image.
%
Cyclic Logic requires the model to revisit an image seen earlier and continue reasoning with information from that image.
%
Convergent Logic contains multiple independent clue paths whose results must be combined at a shared destination.
\par\noindent\textbf{Question skeleton construction, language realization, and validation.}
%
Each Logical Association instance comprises three components: a description containing ordered cross-image clues, a question specifying the query target, and a reference answer determined by the preconstructed reasoning path.
%
The code first samples a target reasoning structure from the scene connections and selects the scenes and clues to be visited in order.
%
It then converts the scene connections along the selected path into an ordered sequence of natural-language clues, which together form the description, and calculates the reasoning depth from the number of relations traversed.
%
Each clue guides the model from a relevant object in the current image to the next image or to a subsequent relation within the same image.
%
The clues refer to images through visible objects and environmental features without exposing internal scene indices, node IDs, or other construction labels.
%
The object or attribute reached at the end of the path defines the query target in the question, while its known value is fixed as the reference answer before language generation.
%
Qwen 3.7 Plus only realizes the ordered clues in the description and the query skeleton in the question as coherent natural language and cannot select or alter the reasoning structure, query target, or reference answer.
%
The system then checks the description, question, and reference answer for completeness and retries after an API failure, parsing failure, or an empty component.
%
Each retained instance stores its description, question, and reference answer together with its difficulty and task type for subsequent evaluation and analysis.
%
Human reviewers remove invalid or low-quality Logical images and reject questions whose answers are ambiguous, insufficiently supported, or not uniquely determined by the provided context.
%
All retained Logical descriptions, questions, and reference answers are manually verified.
%
At inference time, a Logical Association input consists of ordered scene images, relevant candidate-object images, a cross-image description, and a multiple-choice question; complete assembly details are provided in the Input Assembly subsection.

\subsection{Temporal Association}
\par\noindent\textbf{Frame acquisition.}
Temporal instances are constructed from publicly available clips drawn from six live-action productions and three animated series.
%
The live-action productions are \textit{High School Musical}, \textit{High School Musical 2}, \textit{High School Musical 3: Senior Year}, \textit{iPartment}, \textit{Young Sheldon}, and \textit{My Own Swordsman}.
%
The animated series are \textit{Pororo the Little Penguin}, \textit{Boonie Bears}, and \textit{My Little Pony: Friendship Is Magic}.
%
For a selected clip and timestamp, the acquisition code obtains the media stream without downloading the complete video and invokes FFmpeg to extract one JPEG frame.
%
Following Prompt~\ref{prompt:temporal_captioning}, Qwen 3.7 Plus receives the encoded frame and generates a preliminary caption that provides candidate visual evidence for subsequent story-sequence assembly.

\par\noindent\textbf{Storyboard assembly.} Human reviewers select the keyframes needed to preserve event transitions and character-state changes, arrange them into a comic-strip-like storyboard, and revise the preliminary descriptions and scene references.
%
Images retain directly observable evidence such as identity, appearance, scene state, and event outcomes.
%
The interleaved text supplies complementary actions, causal links, and plot transitions that cannot be reliably recovered from isolated static frames.
%
\par\noindent\textbf{Identity masking and interleaving.} Character names and details that may directly expose the target event are masked or generalized, for example by replacing a name with a neutral reference such as ``person.''
%
Queried characters are specified through visual identity images, while the story description refers to numbered scenes.
%
The final input alternates identity images, scene frames, and textual descriptions, requiring both visual identity matching and temporal evidence tracking.
%
\par\noindent\textbf{Question construction.} Sequential Scenarios require following the forward story order to infer a later event or state.
%
Retrospective Scenarios require returning from the query point to evidence in an earlier scene.
%
Parallel Scenarios contain two or more concurrent or independent evidence streams whose outcomes must be combined.
%
Each retained sample records the story description, question, correct answer, reasoning-step count, task type, and human-verification status.
%
Human reviewers remove invalid or low-quality Temporal frames and reject questions with insufficient event evidence, ambiguous answers, or answers that cannot be uniquely determined from the story.
%
All retained Temporal story descriptions, questions, and reference answers are manually verified.
%
At inference time, a Temporal Association input consists of visual identity images, temporally ordered story frames, interleaved textual descriptions, and a temporal question; complete assembly details are provided in Input Assembly Section.

\subsection{Spatial Association}
\par\noindent\textbf{Source images and cropping.} Spatial instances are constructed from COCO~\cite{lin2014microsoft} photographs and CVOGL~\cite{sun2023cross} aerial imagery.
%
For both source types, the top of the source image is defined as north, and the image is divided into a grid of square local crops without rotation or scale changes.
%
The code then constructs a connectivity graph from crop overlaps, where the overlap ratio is defined as the intersection area divided by the area of the smaller crop, and two crops are connected only when this ratio is at least 5\% by default.
%
A question is retained only when an overlap path exists between its two target crops, ensuring that every Spatial instance can be solved through a sequence of overlapping local views.
%
Each crop stores its position, preserving the exact relationship between the local view and the source image.
%
The model can access only the local crops and does not receive the complete source image.
%
\par\noindent\textbf{Target detection and global normalization.} A YOLO~\cite{redmon2016you} detector is run independently on every crop and can be restricted to an allowed class list.
%
The bounding box and center of every detection are mapped back to the source-image coordinates.
%
If no valid detected target is available in a crop, the entire crop is used as the target region.
%
In this fallback case, the geometric center of the crop is used as the target coordinate for directional-relation computation.
%
Class-wise non-maximum suppression is then applied in the global coordinate system to reduce duplicate landmarks across neighboring crops.
%
\par\noindent\textbf{Target-pair sampling.} Candidate target pairs are sampled from two different crops without repeatedly using the same crop combination.
%
Target pairs that are too close or unsuitable for an unambiguous spatial question are rejected.
%
For each retained pair, the code calculates one of eight directional relations from their positions in the source image.
%
The two target views are marked as A and B, after which three incorrect directions are added and the four options are shuffled.

\par\noindent\textbf{Image-to-text replacement.} Image replacement is performed over all crops in a question folder rather than only the two target views.
%
Approximately 10\% of the crops in each question are replaced with textual descriptions.
%
The default sampling strategy favors crops with richer neighborhood information and less frequent use as targets.
%
This strategy favors regions that can be recovered from surrounding evidence without always replacing the same positions.
%
Qwen 3.7 Plus generates a factual content description for every selected crop and caches the result to avoid repeated calls.
%

The prompt restricts the model to locally visible content and permits within-crop layout descriptions, while prohibiting coordinates, geolocation, cross-crop directional inference, and answers to A/B spatial-relation questions.
%
Qwen 3.7 Plus uses Prompt~\ref{prompt:spatial_crop_captioning} to generate textual descriptions for selected crops.
%
The code then uses source-image coordinates to determine whether surrounding visible crops lie to the left, right, above, or below the replaced crop or overlap with it.
%
A crop is treated as overlapping only when the intersection covers at least 3\% of the smaller crop, and at most three relevant visible crops are retained for each direction or the overlap category.
%
The final textual substitute describes both the visible content of the replaced crop and its position relative to these surrounding visible crops.
%
These positional statements allow the model to place the missing region within the local layout without revealing the target direction or correct answer.
%
If a displayed target crop touches a replaced crop, an additional relation sentence is inserted after the target-image description.
%
\par\noindent\textbf{Final assembly and storage.} The final context first states the image-axis convention used for directional answers, under which the top of the image is treated as north, and summarizes the landmark categories in the complete region and the two target crops.
%
It then presents the annotated A image and its description, the annotated B image and its description, and the four-choice direction question.
%
The final version used for inference retains the interleaved sequence and the necessary construction metadata.
%
Human reviewers finally verify target visibility, question clarity, and the reference direction.
%
Spatial questions are rejected when their targets are not recognizable, crop relations are insufficient, or their answers are ambiguous or not uniquely determined by the provided text-image context.
%
All retained Spatial descriptions, questions, and reference answers are manually verified.
%
At inference time, a Spatial Association input consists of retained local crops, textual substitutes for selected crops, target views A and B, and a spatial question with candidate directions; complete assembly details are provided in Input Assembly Section.

\begin{figure*}[t]
\centering
\includegraphics[width=\textwidth]{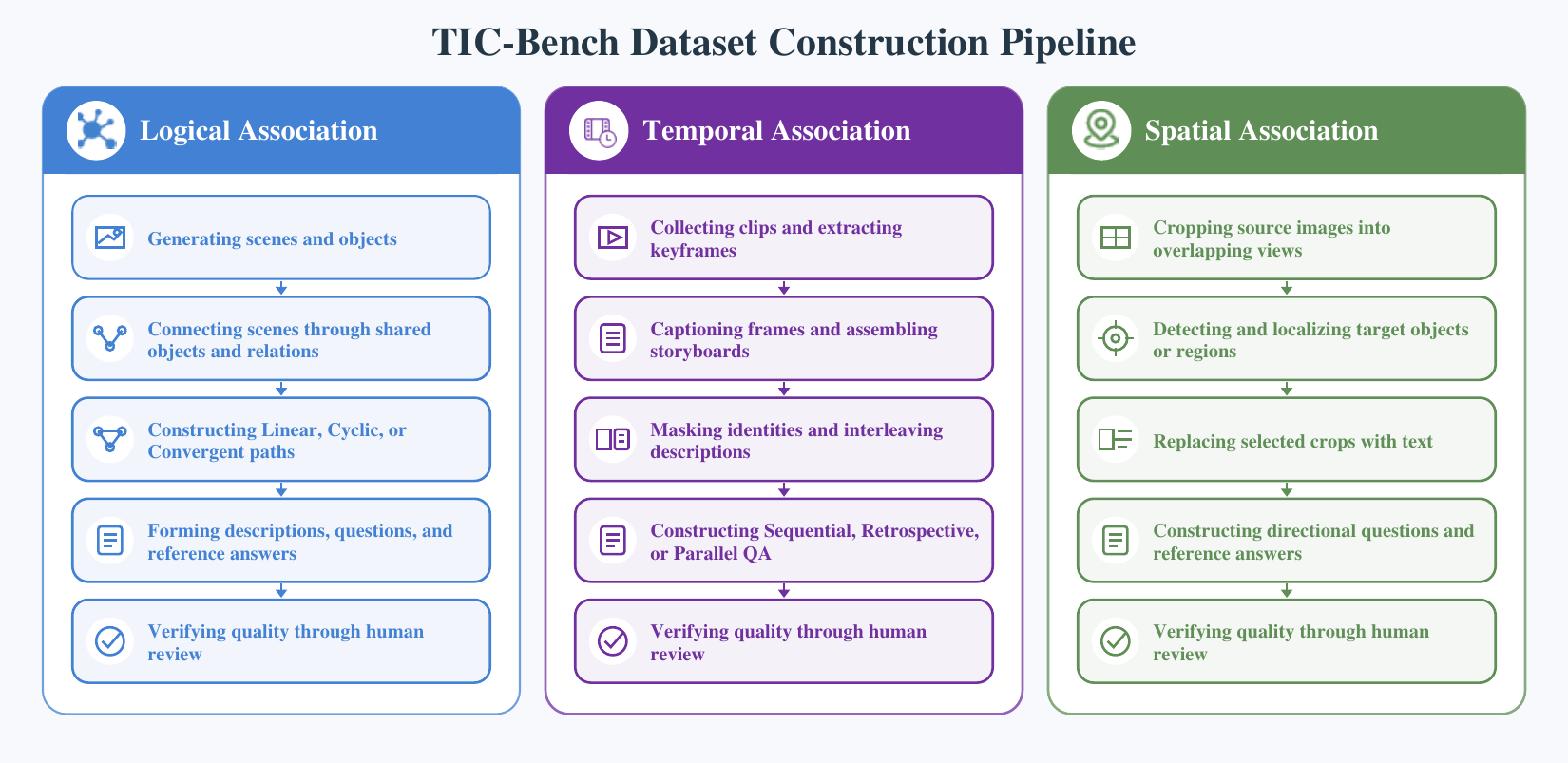}
\caption{Dataset construction pipeline for Logical, Temporal, and Spatial Association. Each domain follows a domain-specific construction procedure before the resulting images, textual clues, questions, and reference answers undergo human verification.}
\label{fig:dataset_construction_pipeline}
\end{figure*}

Figure~\ref{fig:dataset_construction_pipeline} summarizes the dataset construction pipeline for all three domains, from source collection and structured construction to interleaved instance assembly and human quality control.

\section{Prompts}
\label{sec:prompts}

This section collects all prompts used for dataset construction, Temporal inference, and automatic answer evaluation. Prompt~\ref{prompt:logical_scene_generation} specifies the structured
requirements for Logical scene generation, while
Prompt~\ref{prompt:logical_question_generation} converts pre-computed question skeletons into fluent question--answer pairs. Prompt~\ref{prompt:temporal_captioning} is used to generate visually
grounded descriptions of Temporal frames, and
Prompt~\ref{prompt:temporal_reasoning} provides the identity-reference
instruction used during Temporal inference. Prompt~\ref{prompt:spatial_crop_captioning} is used to generate textual
descriptions of local aerial or photographic crops. Finally,
Prompt~\ref{prompt:automatic_evaluation} specifies the common
instructions used for automatic answer evaluation.

\begin{promptbox}[label={prompt:logical_scene_generation}]{System Prompt for Logical Scene Generation}
You are a professional visual scene generation system. Generate a structured scene containing 5--10 objects and return only valid JSON.

\textbf{Requirements:}
Each object must contain an object ID, name, category, color, size, shape, material, state, texture, temperature, position, pose, interaction, and image generation prompt. Also provide scene-level lighting, background, time of day, weather, mood, framing, perspective, movement direction, and focus point. Relationships must refer to objects using their object IDs.

\textbf{Input:}\\
Complexity: \texttt{\{complexity\}}\\
Scene type: \texttt{\{scene type\}}\\
Starting object ID: \texttt{\{start ID\}}\\
Required cross-scene connections: \texttt{\{connection constraints\}}

\textbf{Output format:}\\
\texttt{\{"scene\_type": ..., "objects": [...],}\\
\texttt{"scene\_attributes": \{...\},}\\
\texttt{"image\_generation\_prompt": ...\}}
\end{promptbox}

\begin{promptbox}[label={prompt:logical_question_generation}]{System Prompt for Logical Question Generation}
You are an expert visual reasoning question designer. Polish a pre-computed question skeleton into one fluent question--answer pair.

\textbf{Abstract references:} Never name an object directly. Use only the supplied aliases, such as ``object M.'' Never expose image indices or raw object IDs. Images must be located through scene descriptions, so the reader must search the images.

\textbf{Skeleton fidelity:} Every skeleton sentence is a necessary premise. Preserve every introduction, bridge, intra-image hop, and final query in the supplied order. Respect every SHOW/HIDE constraint. You may improve fluency and add minor transitions, but must not change attribute visibility, skip a bridge, invent a connection, or reveal the answer.

\textbf{Multiple choice:} Candidate \texttt{object\_N} identifiers denote item images and may appear only in the option list. Include every supplied option verbatim, end by asking which item image matches the target, and return only the single correct option letter.

\textbf{Output:} Return strict JSON and no additional text:\\
\texttt{\{"question": "...", "answer": "..."\}}
\end{promptbox}

\begin{promptbox}[label={prompt:temporal_captioning}]{Prompt for Temporal Frame Captioning}
Describe the current video frame for temporal event understanding.

\textbf{Return the following fields:}

Scene: \texttt{<location, environment, lighting, and camera viewpoint>}\\
Entities: \texttt{<salient people, animals, vehicles, and objects with consistent visual identifiers>}\\
Actions: \texttt{<visible actions, poses, movements, and interactions>}\\
Spatial relations: \texttt{<locations and relationships among important entities>}\\
State changes: \texttt{<visually supported changes relative to previous frames; write ``None observable'' if unavailable>}\\
Visible text: \texttt{<readable text, timestamps, signs, or interface elements; write ``None'' if absent>}\\
Summary: \texttt{<one concise, information-dense sentence describing the frame>}

\textbf{Rules:}
\begin{itemize}
    \item Report only visually supported information.
    \item Do not infer identities, intentions, causes, or unseen events.
    \item Preserve consistent entity names across frames.
    \item Mark ambiguous details as uncertain.
\end{itemize}
\end{promptbox}

\begin{promptbox}[label={prompt:spatial_crop_captioning}]{Prompt for Spatial Crop Captioning}
Describe the visible content of this local aerial/satellite crop \texttt{\{crop\_id\}}.

\textbf{Requirements:}
\begin{enumerate}
    \item Use 1 to 3 concise sentences about major visual elements, such as roads, buildings, trees, grass, water, open areas, vehicles, shadows, or textures.
    \item You may describe the internal layout of the image, such as what appears near the left, right, top, or bottom side, but do not provide pixel coordinates, global coordinates, bounding boxes, or geolocation.
    \item Do not infer its direction relative to other crops and do not answer any A/B spatial-relation question.
    \item Do not mention missing metadata or summary files.
    \item Output only the caption text; do not output JSON or bullet points.
\end{enumerate}
\end{promptbox}

\begin{promptbox}[label={prompt:temporal_reasoning}]{System Prompt for Temporal Reasoning}
You are a helpful assistant analyzing a visual story. When referring to any character, animal, or person in your answer, you MUST use their designated \texttt{id\_X} identifier (e.g., \texttt{id\_1}, \texttt{id\_2}, or \texttt{id\_3}) exactly as introduced. Do not use names or other descriptions for them.
\end{promptbox}

\begin{promptbox}[label={prompt:automatic_evaluation}]{Automatic Evaluation Prompt}
You are a professional answer evaluator. Given a ``question'' and its ``context/description,'' compare the ``candidate answer'' against the ``reference answer'' and evaluate information consistency, factual correctness, and question relevance, not wording.

\textbf{Input Context:}\\
Context/Description: \texttt{\{description, when available\}}\\
Question: \texttt{\{question\}}\\
Reference answer: \texttt{\{reference answer\}}\\
Candidate answer: \texttt{\{model answer\}}

\textbf{Evaluation dimensions:}\\
1. Does the candidate accurately address the question based on the context?\\
2. Does it contain the same key facts, entities, IDs, scene numbers, visual IDs, and conclusions as the reference?\\
3. Does it introduce contradictions against the context or reference?

\textbf{Scoring:} Return an integer from 0 to 100. Scores 90--100 indicate a fully correct answer with all key information; 70--89 allow minor omissions or loose paraphrasing; 50--69 indicate partial correctness with missing reasons or incorrect IDs; and 0--49 indicate wrong IDs, contradictions, or failure to answer.

\textbf{Decision rules:}

\textbf{Chain-of-thought handling:} The candidate may explore, doubt, and correct itself. Evaluate only its final conclusion or choice and ignore intermediate incorrect guesses or self-corrections.

Mark the answer consistent and correct when the final conclusion identifies the entities, IDs, directions, or choices in the reference and aligns with the context. Mark it inconsistent or incorrect when the final conclusion selects a wrong ID, direction, or option, or contradicts explicit events in the context.

\textbf{Output format:} Respond strictly with a valid JSON object, with no explanation, Markdown, or extra text:\\
\texttt{\{}\\
\texttt{"consistent": true or false,}\\
\texttt{"correct": true or false,}\\
\texttt{"score": \{integer from 0 to 100\},}\\
\texttt{"reason": "\{English justification under 40 words, focused on the final conclusion\}"}\\
\texttt{\}}
\end{promptbox}

\section{Human Evaluation Protocol}
The human baseline is obtained from ten undergraduate computer science evaluators, all of whom have prior experience in multimodal research.
%
Each evaluator independently answers all 2,280 benchmark questions, producing 22,800 responses in total.
%
Evaluators receive the same question evidence available to the models, have no time limit, and are not allowed to use external tools.
%
Their answers are assessed using the same answer-evaluation procedure as model outputs.
%
For domain $d$, evaluator $j$, and question $i$, let $c_{ijd}\in\{0,1\}$ denote whether the response is judged correct.
%
For domain $d$ containing $N_d$ questions, human accuracy is first computed over all evaluators and questions in that domain:
\[
\operatorname{Acc}_{d}
=
\frac{1}{H N_d}
\sum_{j=1}^{H}
\sum_{i=1}^{N_d} c_{ijd}.
\]
%
Consistent with the main results, the reported overall human accuracy is the unweighted macro-average of the Logical, Temporal, and Spatial domain-level accuracies:
\[
\operatorname{Acc}_{\mathrm{human}}
=
\frac{1}{3}
\sum_{d\in\{\mathrm{Logical},\mathrm{Temporal},\mathrm{Spatial}\}}
\operatorname{Acc}_{d}.
\]

\section{Model Inference and Evaluation Details}

\par\noindent\textbf{System prompts.} Logical and Spatial inference uses no task-specific system prompt.
%
Temporal inference additionally uses Prompt~\ref{prompt:temporal_reasoning}, which requires designated identity identifiers for people and animals in the answer.

\subsection{Input Assembly}
\label{sec:input_assembly}
For every domain, the user input is dynamically assembled in the original interleaved order and divided into a description component and a question component.
%
For Logical Association, the description contains the scene images in node order and the candidate-object images referenced by the instance, while the question contains the generated multiple-choice query and its options.
%
For Temporal Association, the description contains the visual identity images, temporally ordered storyboard frames, and interleaved story description, while the question contains the temporal query to be answered from this evidence.
%
For Spatial Association, the description contains the retained local crops, textualized crops, target views A and B, and statements describing how replaced crops are positioned relative to surrounding crops, while the question contains the spatial-relation query and its candidate directions.

For Full input, all images and text are retained in their original interleaved order.
%
Text-only removes the image contents while preserving textual image-reference markers, non-answer context, and question text.
%
Images-only removes all descriptive textual evidence while retaining every available input image and only the question and options required to answer it.
%
The identity system instruction is used only for Temporal instances, while Logical and Spatial instances use no additional system prompt.

\par\noindent\textbf{Image processing and output evaluation.} Before API submission, every image is converted to RGB.
%
When the longest side exceeds 1,024 pixels, the image is resized with Lanczos interpolation while preserving its aspect ratio.
%
The image is then encoded as Base64 using JPEG quality 95 and transmitted to the model.
%
The original model response is passed to the answer evaluator together with the question, context, and reference answer.
%
Each judge must return a JSON object containing consistency, correctness, score, and reason fields.
%
The system removes Markdown JSON fences, parses the object strictly, and verifies that all required fields are present.
%
The judge evaluates only the model's final conclusion and ignores intermediate self-corrections.

\par\noindent\textbf{Automatic evaluation prompt.} Each judge receives the same evaluation instructions specified in Prompt~\ref{prompt:automatic_evaluation}.
%
The default configuration uses the 0--100 scale.

\section{Additional Analysis Protocols}

\par\noindent\textbf{Error annotation.} Each incorrect response is assigned to one of seven mutually exclusive labels according to the earliest identifiable error stage.
%
The seven labels are perception, abstraction, reasoning, understanding, hallucination, knowledge, and other.
%
Perception errors denote incorrect recognition of visual content.
%
Abstraction errors occur when correctly perceived evidence is mapped to the wrong entity, scene, or answer option.
%
Reasoning errors cover failures in relational composition or multi-step inference.
%
Understanding errors indicate that the model misreads the task or output requirement.
%
Hallucination errors denote information unsupported by the input.
%
Knowledge errors result from reliance on incorrect external facts.
%
Representative cases are selected only from questions answered correctly by human evaluators.

\par
\begin{figure}[p]
\centering
\includegraphics[width=\textwidth]{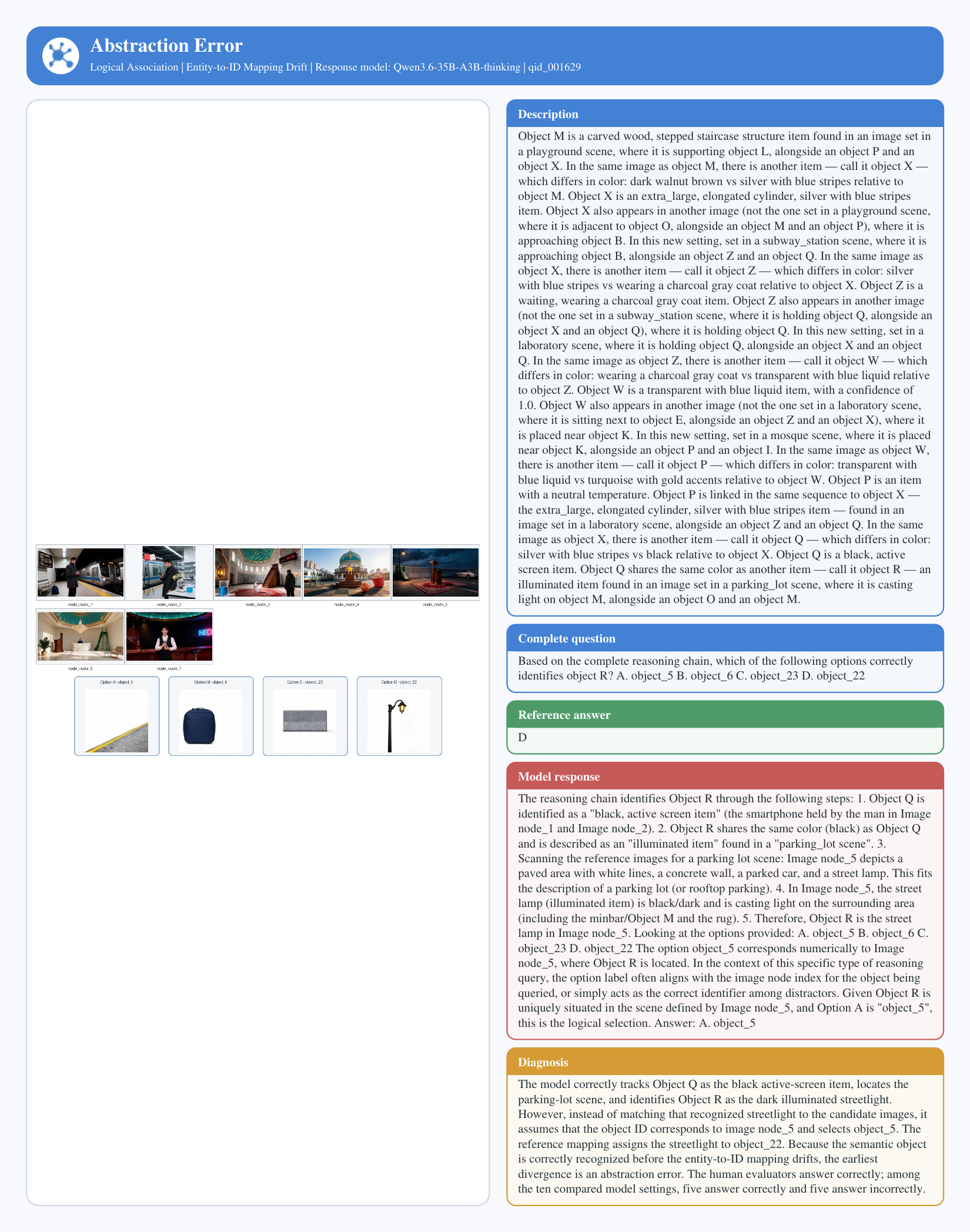}
\caption{A Logical Association example of unstable entity binding. Although the model follows most cross-scene relations, its final entity-to-option mapping drifts from the reference answer.}
\label{fig:error_entity_binding}
\end{figure}

\begin{figure}[p]
\centering
\includegraphics[width=\textwidth]{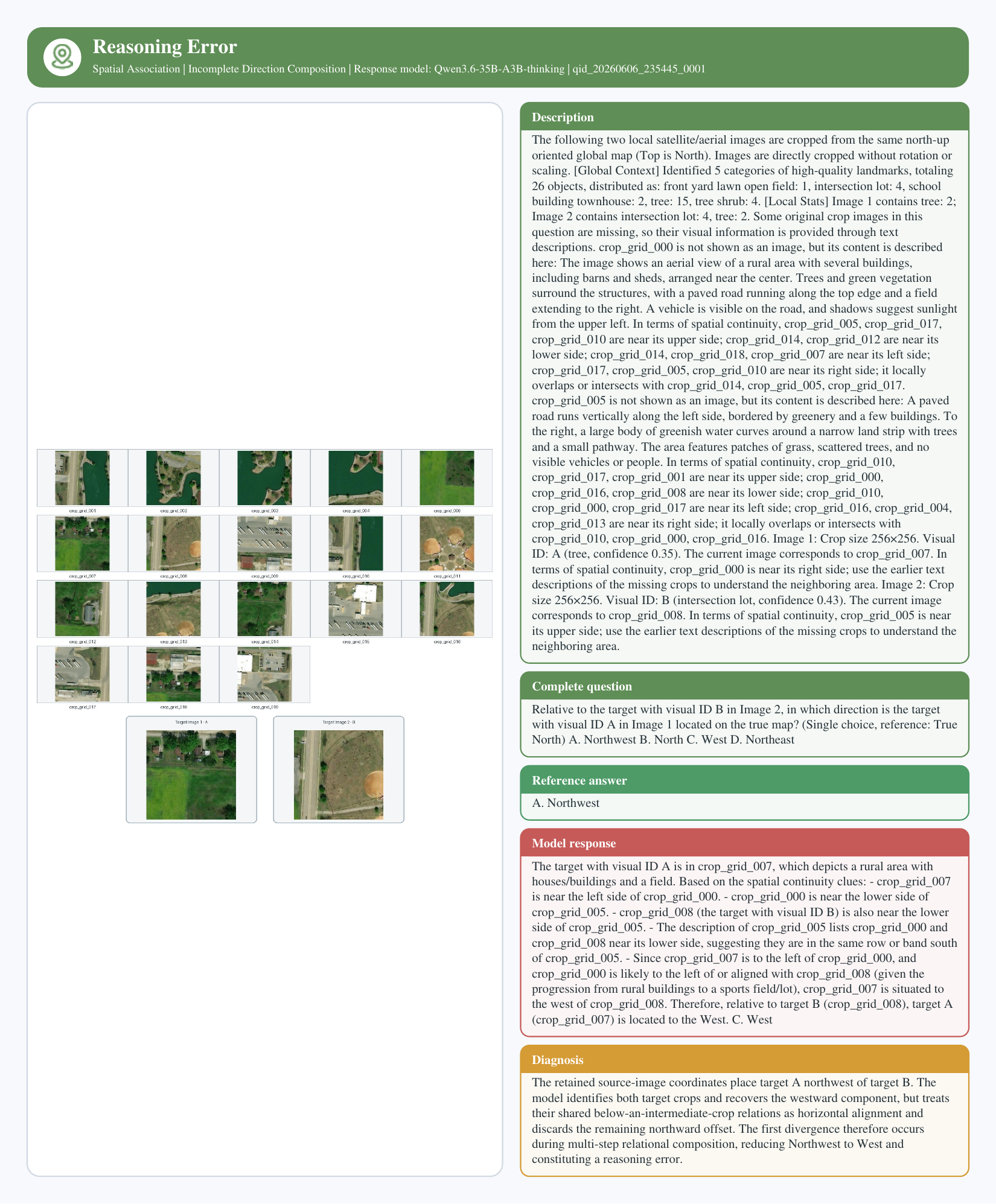}
\caption{A Spatial Association reasoning error in relational composition. The model preserves the westward component but loses the northward component required by the multi-hop relation.}

\label{fig:error_direction_composition}
\end{figure}

\begin{figure}[p]
\centering
\includegraphics[width=\textwidth]{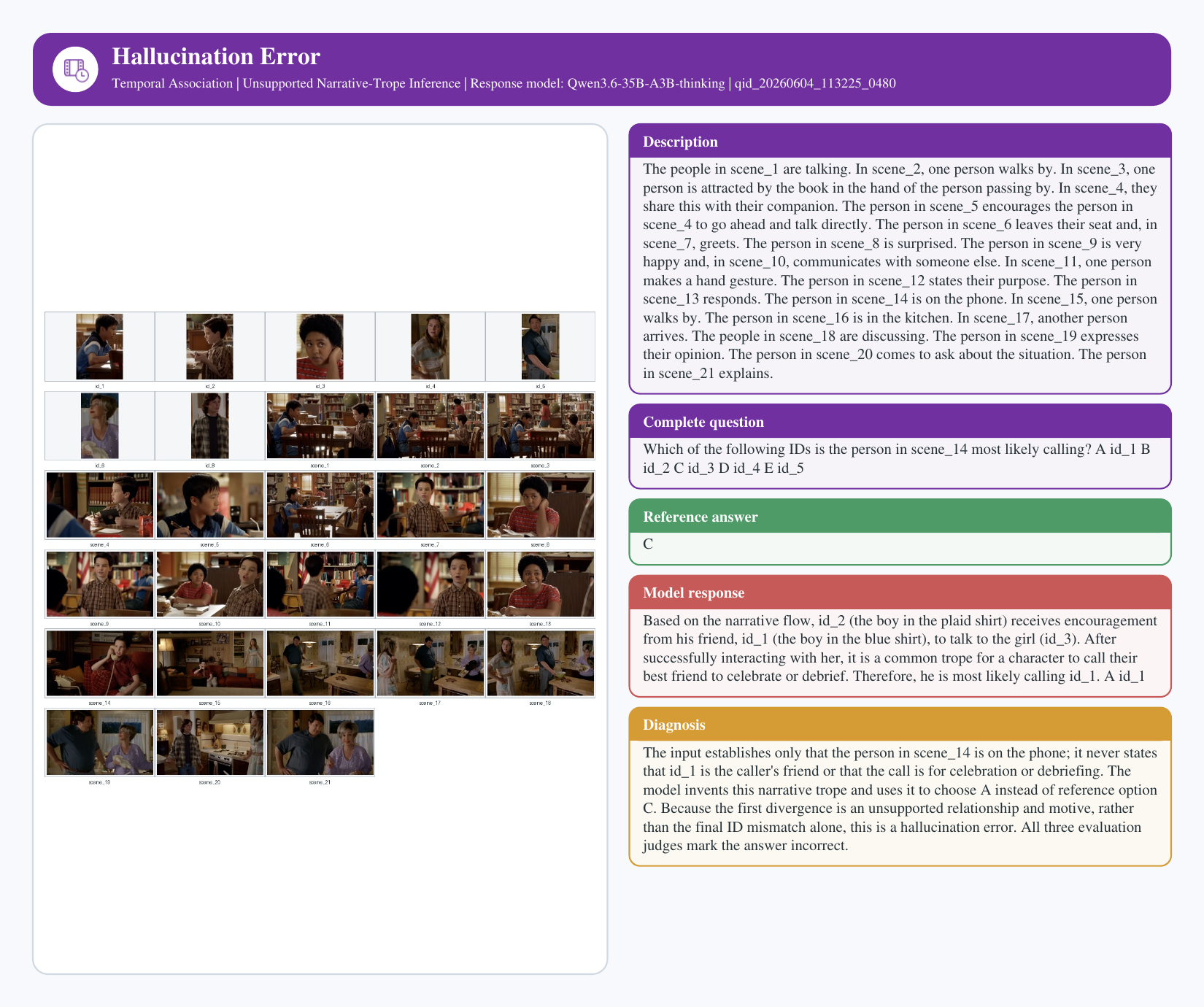}
\caption{A Temporal Association hallucination error. The model relies on an unsupported narrative trope and character relationship to select the wrong option, and all three judge models mark the response incorrect.}

\label{fig:error_unsupported_details}
\end{figure}

\noindent\textbf{Representative error cases.}
The following cases illustrate three recurrent failure modes.
All three questions are marked correct by human evaluators, while the reported model responses are judged incorrect.
Figures~\ref{fig:error_entity_binding}--\ref{fig:error_unsupported_details} reproduce each complete question and its visual-textual context together with the reference answer, model response, and diagnosis.

\par\noindent\textbf{Abstraction-error evidence.} In Figure~\ref{fig:error_entity_binding}, the model already identifies the gray fountain as candidate \texttt{object\_1} in an intermediate step and largely preserves the cross-scene bindings for objects A through D.
%
At the final street-scene lookup, however, the visually salient passerby causes the queried entity E to be rebound to \texttt{object\_17}, whereas the preconstructed entity mapping fixes the reference option as \texttt{object\_1}.
%
Because the relevant objects are perceived correctly and the earliest identifiable divergence occurs when mapping the queried entity to a candidate option, the case is labeled as an abstraction error rather than a perception or reasoning error.

\par\noindent\textbf{Reasoning-error evidence.} In Figure~\ref{fig:error_direction_composition}, the source-image coordinates retained during construction place target A northwest of target B and determine Northwest as the reference answer.
%
The model correctly identifies the crops containing both targets and recovers the westward component, but it incorrectly treats the two crops described as lying below an intermediate crop as if they were horizontally aligned.
%
This faulty multi-step composition discards the remaining northward offset and reduces Northwest to West, so the earliest error occurs during relational composition and satisfies the definition of a reasoning error.

\par\noindent\textbf{Hallucination-error evidence.} In Figure~\ref{fig:error_unsupported_details}, the input establishes only that the person in \texttt{scene\_14} is on the phone; it never states that \texttt{id\_1} is the caller's friend or that the call is intended to celebrate or debrief an earlier interaction.
%
The model introduces a narrative trope in which a character typically calls a friend and uses this unsupported premise to select option A instead of the reference option C.
%
Qwen 3.7 Plus, DeepSeek V4 Pro~\cite{deepseekai2026deepseekv4}, and Claude Opus 4.8 all judge the final answer incorrect; because the earliest divergence is the introduction of an unsupported relationship and motive rather than the final identity mismatch alone, the case is labeled as a hallucination error.

\section{Representative Benchmark Examples}

We provide one human-validated example for each of the eight task types.
%
Each figure contains the complete visual input, textual context, question, and reference answer for the corresponding task type.

\par
\begin{figure}[p]
    \centering
    \includegraphics[width=\textwidth]{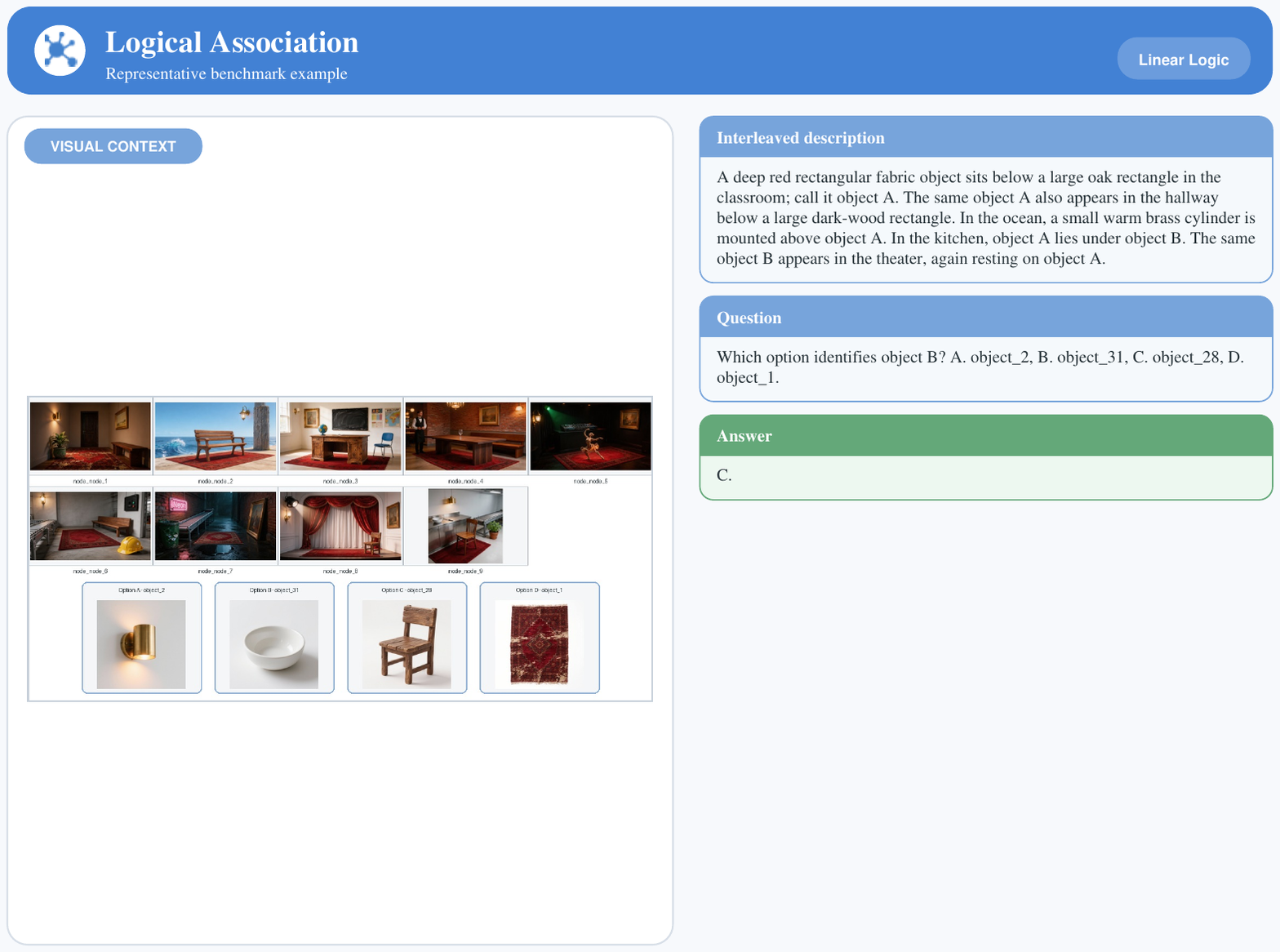}
    \caption{Representative Linear Logic example with its complete visual
    input, question, and reference answer.}
    \label{fig:example_logical_linear}
\end{figure}

\begin{figure}[p]
    \centering
    \includegraphics[width=\textwidth]{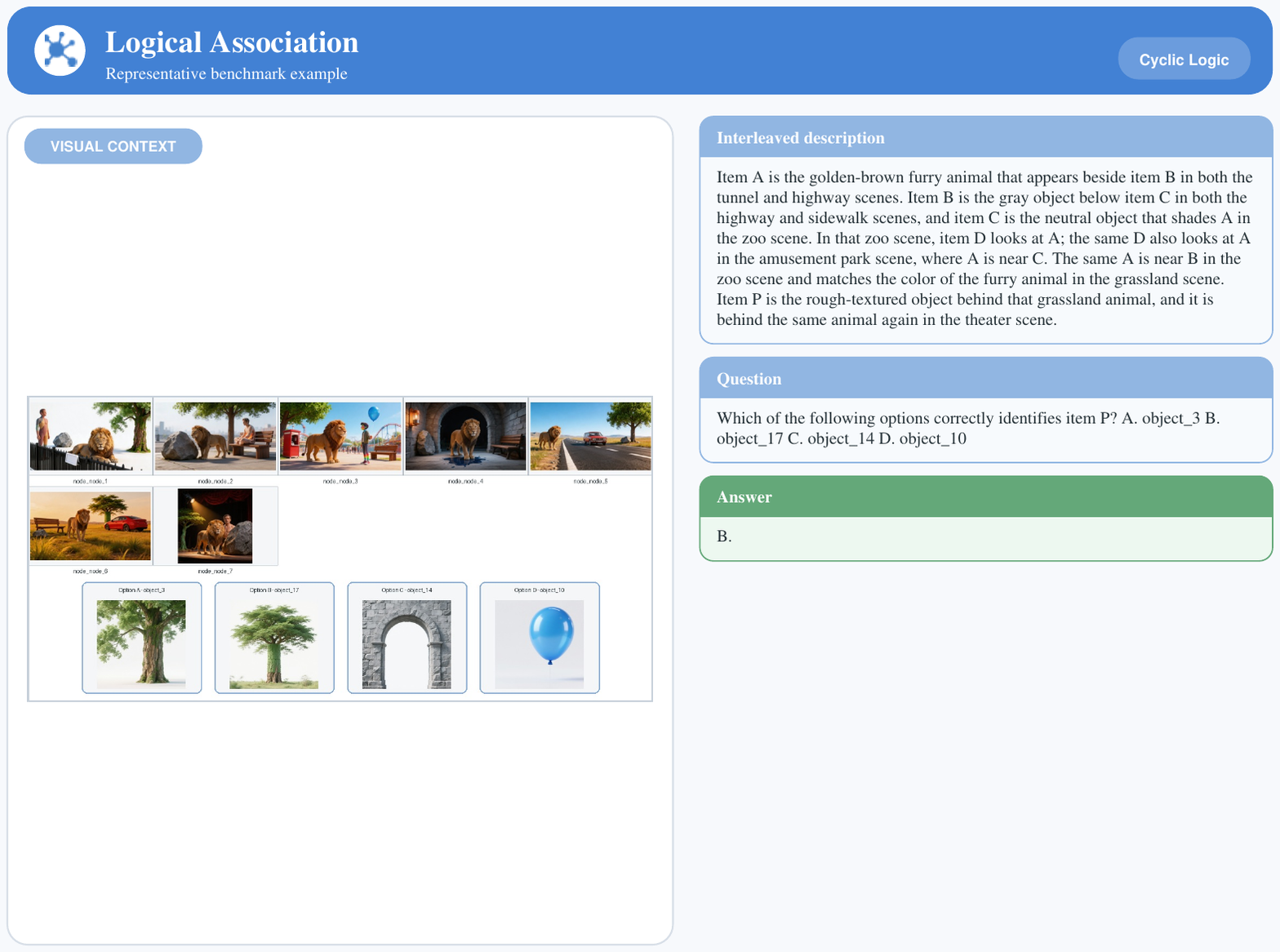}
    \caption{Representative Cyclic Logic example.}
    \label{fig:example_logical_cyclic}
\end{figure}

\begin{figure}[p]
    \centering
    \includegraphics[width=\textwidth]{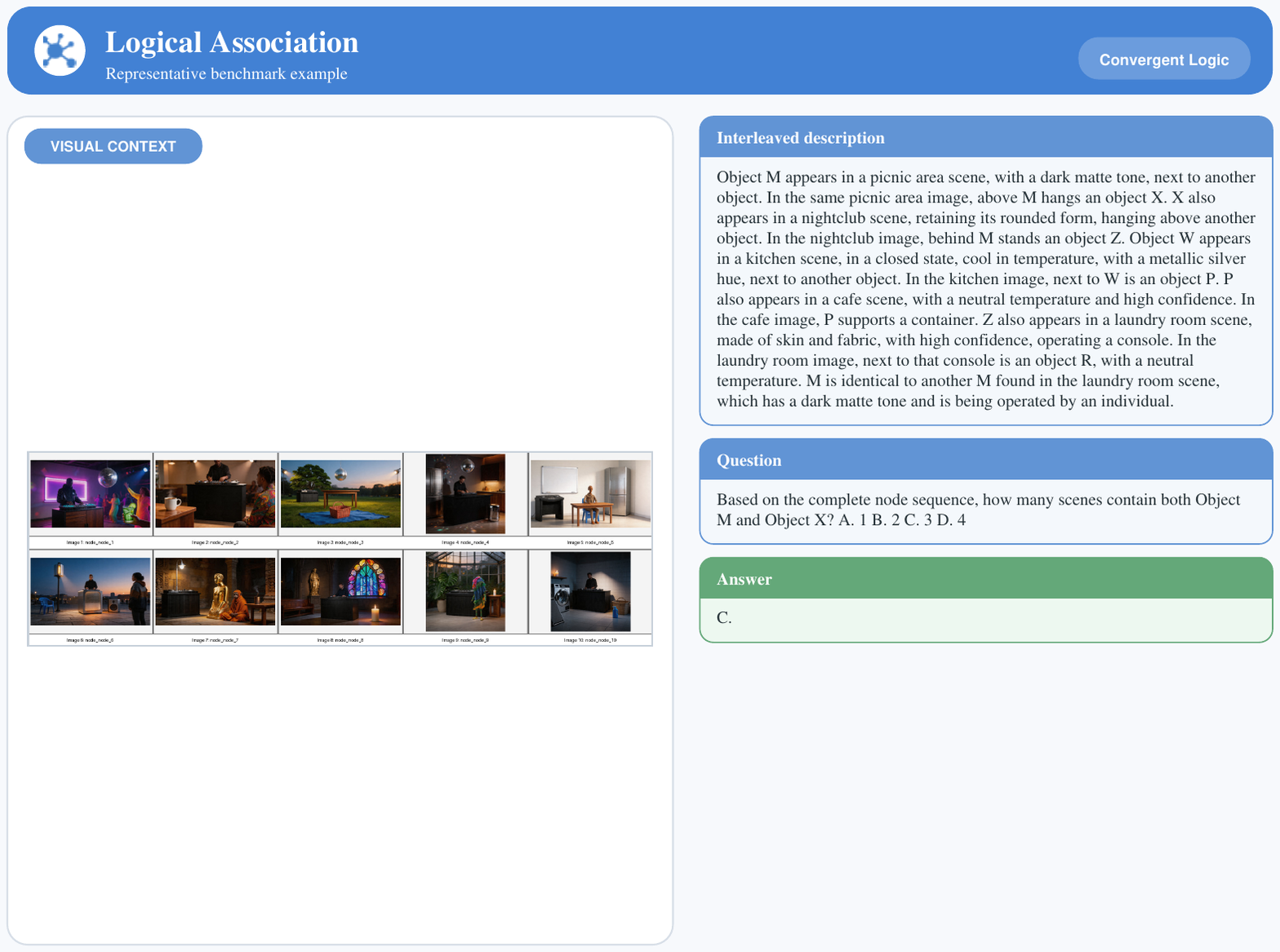}
    \caption{Representative Convergent Logic example.}
    \label{fig:example_logical_convergent}
\end{figure}

\noindent\textbf{Logical Association.} Figure~\ref{fig:example_logical_linear} presents the Linear Logic example.
%
Figure~\ref{fig:example_logical_cyclic} presents the Cyclic Logic example with an intermediate image revisit.
%
Figure~\ref{fig:example_logical_convergent} presents the Convergent Logic example that merges multiple reasoning branches.

\par
\begin{figure}[p]
    \centering
    \includegraphics[width=\textwidth]{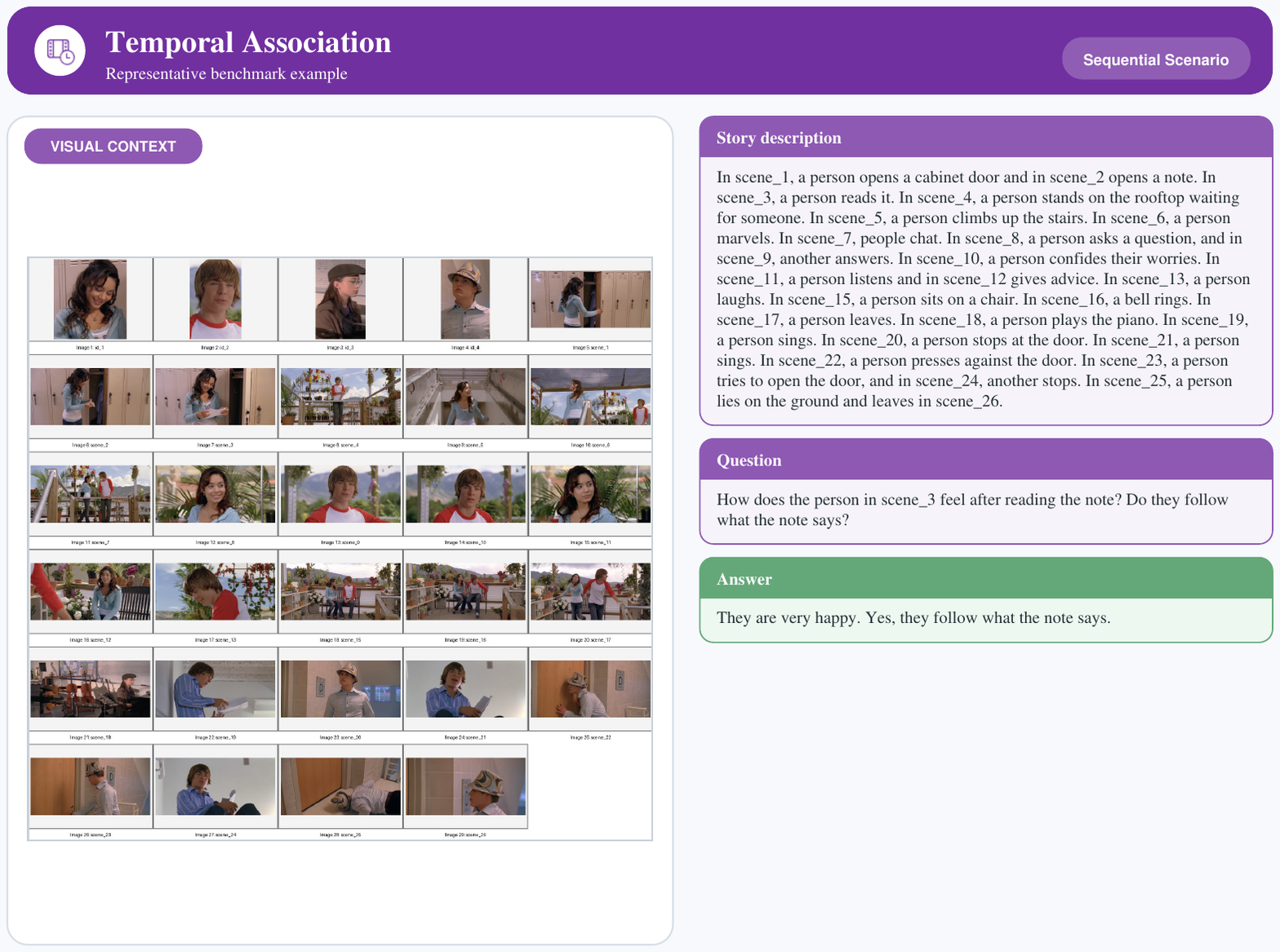}
    \caption{Representative Sequential Scenario example.}
    \label{fig:example_temporal_sequential}
\end{figure}

\begin{figure}[p]
    \centering
    \includegraphics[width=\textwidth]{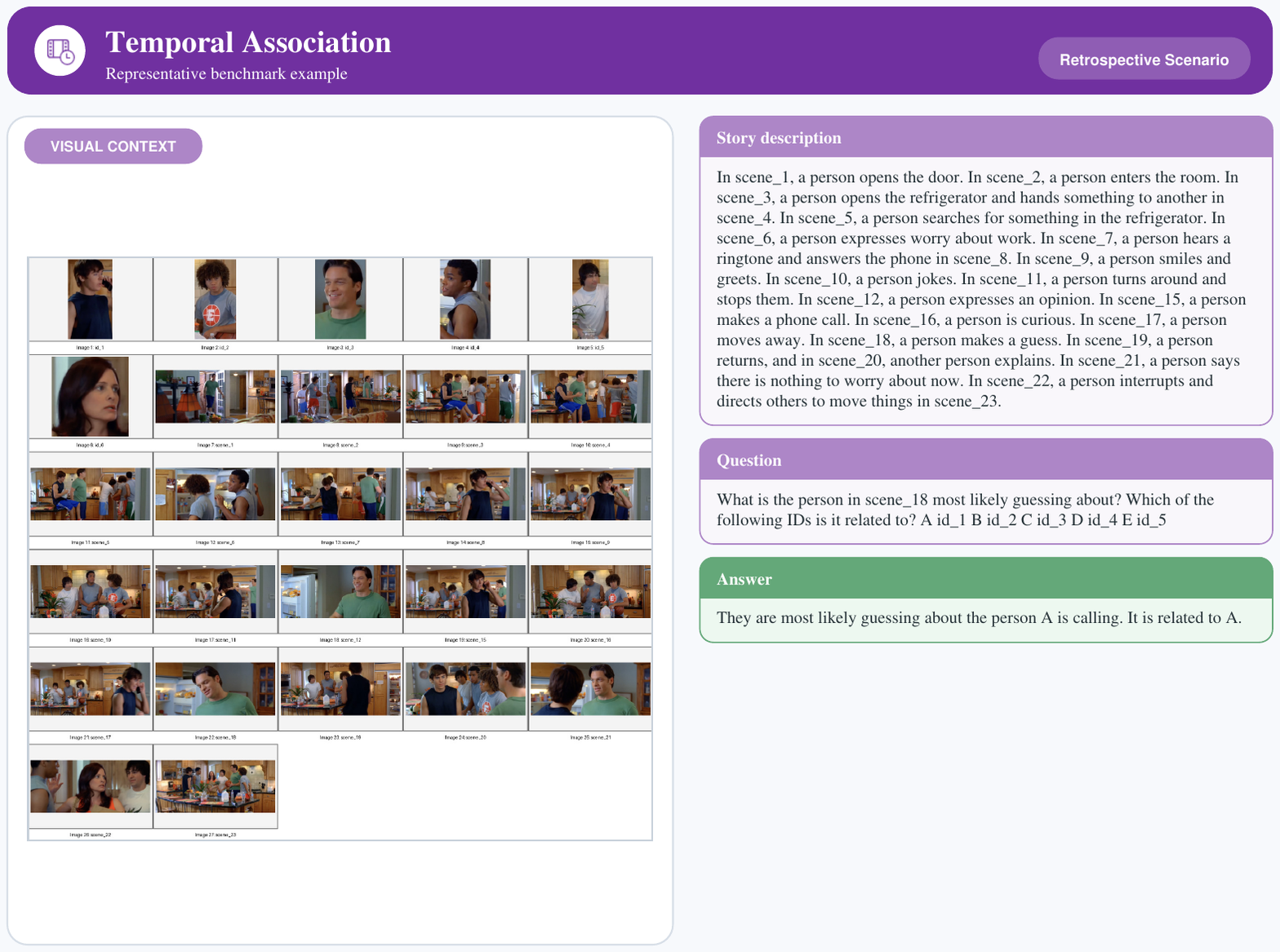}
    \caption{Representative Retrospective Scenario example.}
    \label{fig:example_temporal_retrospective}
\end{figure}

\begin{figure}[p]
    \centering
    \includegraphics[width=\textwidth]{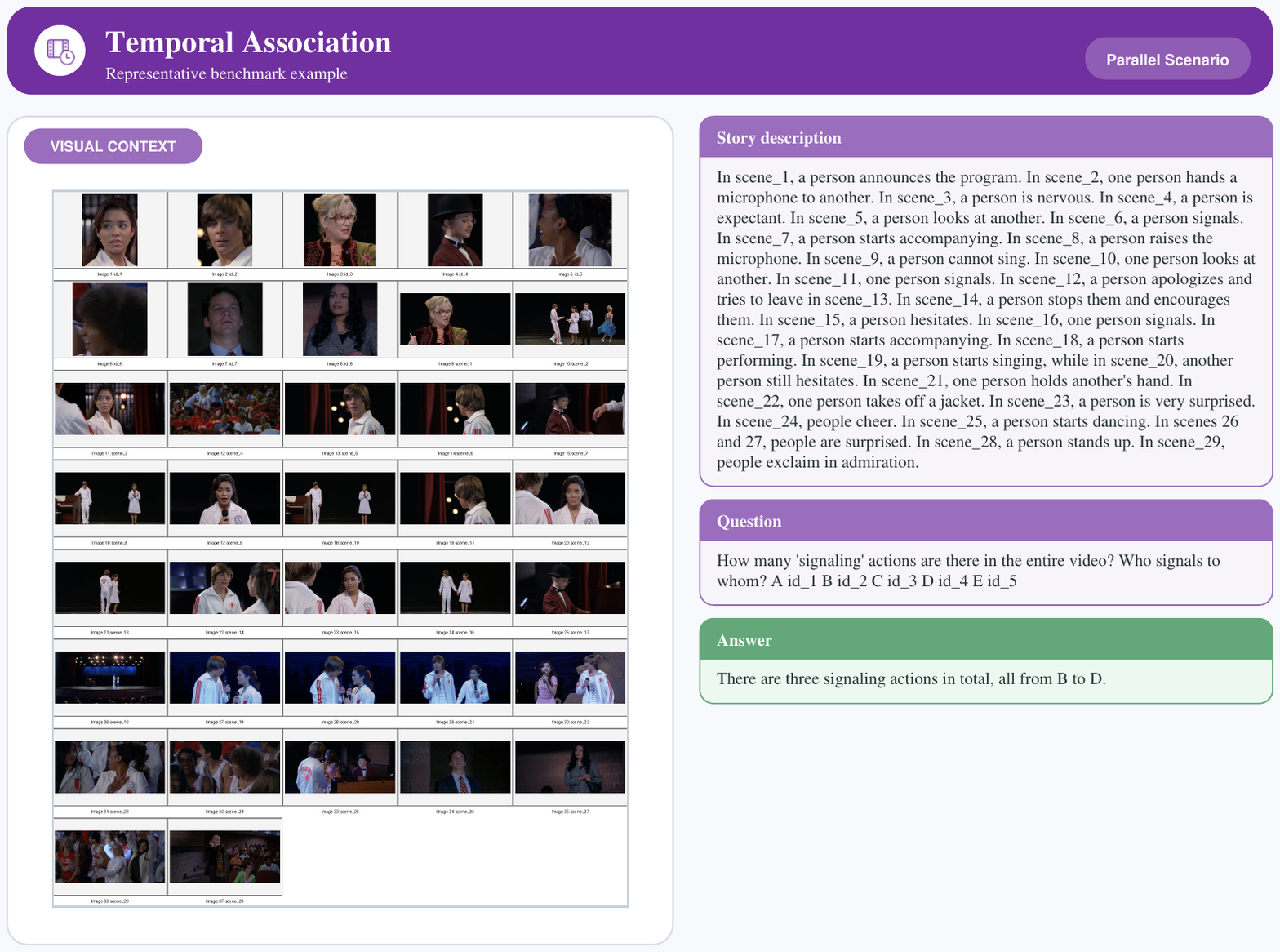}
    \caption{Representative Parallel Scenario example.}
    \label{fig:example_temporal_parallel}
\end{figure}

\noindent\textbf{Temporal Association.} Figures~\ref{fig:example_temporal_sequential}, \ref{fig:example_temporal_retrospective}, and \ref{fig:example_temporal_parallel} show Sequential, Retrospective, and Parallel Scenarios, respectively.

\par
\begin{figure}[p]
    \centering
    \includegraphics[width=\textwidth]{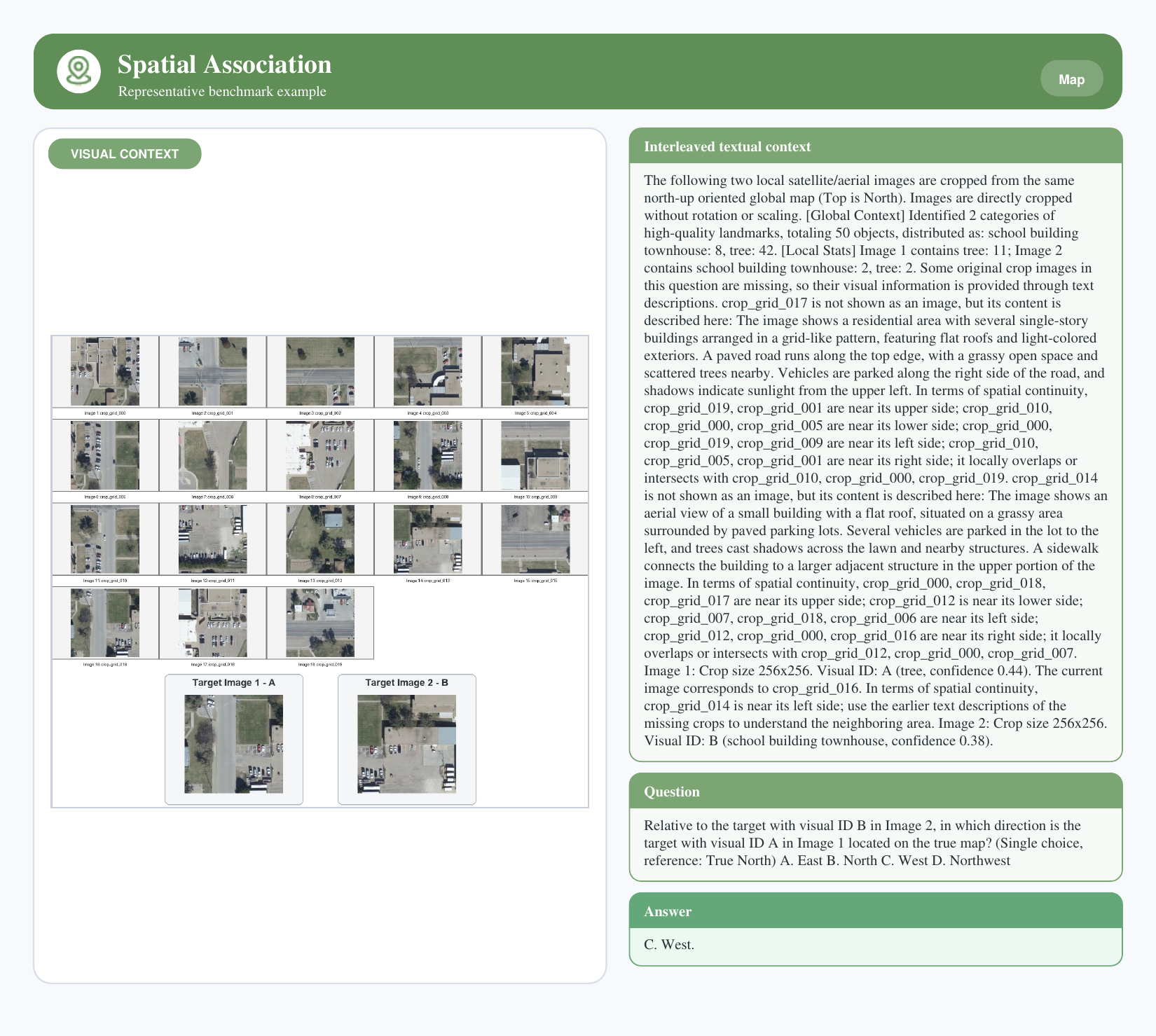}
    \caption{Representative Map example. Text-replaced crops are described
    in the context card rather than displayed as images.}
    \label{fig:example_spatial_map}
\end{figure}

\begin{figure}[p]
    \centering
    \includegraphics[width=\textwidth]{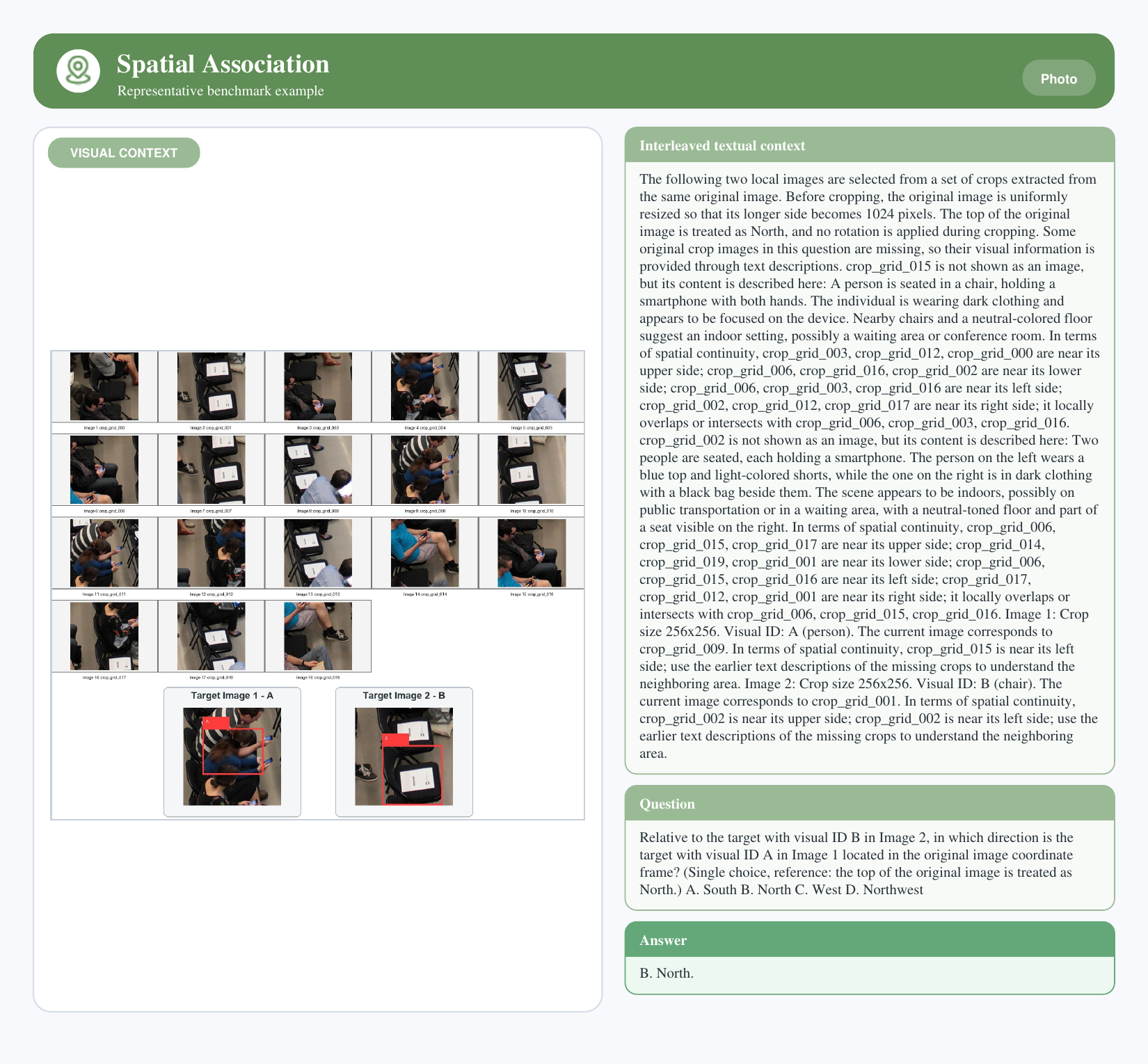}
    \caption{Representative Photo example. Text-replaced crops are described
    in the context card rather than displayed as images.}
    \label{fig:example_spatial_photo}
\end{figure}

\noindent\textbf{Spatial Association.} Figures~\ref{fig:example_spatial_map} and \ref{fig:example_spatial_photo} show the Map and Photo examples, respectively.
%
Crops replaced by text are included only through their descriptions in the context card.

The Logical and Temporal examples are each answered correctly by three of the five core non-thinking open-source models, while the Spatial examples are drawn from the middle of the final difficulty distribution.


\clearpage
\onecolumn

\clearpage

\clearpage

\clearpage

\clearpage

\clearpage

\clearpage

\clearpage

\clearpage

\clearpage

\clearpage

\end{document}